\documentclass{article}
\usepackage{iclr2027_conference,times}

\usepackage{amsmath,amsfonts,bm,amsthm,amssymb}
\usepackage{hyperref}
\usepackage{url}
\usepackage{graphicx}
\usepackage{xcolor}
\usepackage{xurl}
\usepackage[capitalize,noabbrev]{cleveref}

\makeatletter
\iclrfinalcopy
\author{%
    \begin{tabular*}{\textwidth}{@{\extracolsep{\fill}}lll@{}}
      \normalsize\bfseries Haodong Wen\thanks{Equal contribution.
        Emails: \texttt{whd25@mails.tsinghua.edu.cn} and
        \texttt{kaiyuew@stanford.edu}.} &
      \normalsize\bfseries Kaiyue Wen\textsuperscript{*} &
      \normalsize\bfseries Jiaye Teng\thanks{Corresponding author.
        Email: \texttt{tengjiaye@sufe.edu.cn}.}\\[2pt]
      Tsinghua University & Stanford University & Shanghai University of\\
      & & Finance and Economics
    \end{tabular*}%
}
\hypersetup{
    hidelinks,
    pdfauthor={Haodong Wen, Kaiyue Wen, Jiaye Teng},
    pdftitle={Tight Transition Time Bounds for Separable Logistic Regression at the Edge of Stability}
}
\renewcommand{\@maketitle}{%
    \vbox{\hsize\textwidth
      {\LARGE\scshape\raggedright\hyphenpenalty=10000 \@title\par}%
      \vskip 0.22in
      {\centering\normalfont\@author\par}%
      \vskip 0.22in
    }%
}
\makeatother
\renewcommand{\headrulewidth}{0pt}
\crefname{appendix}{Appendix}{Appendices}
\Crefname{appendix}{Appendix}{Appendices}
\crefname{subappendix}{Appendix}{Appendices}
\Crefname{subappendix}{Appendix}{Appendices}

\AddToHook{cmd/appendix/after}{%
  \crefalias{section}{appendix}%
  \crefalias{subsection}{subappendix}%
  \crefalias{subsubsection}{subsubappendix}%
}

 \theoremstyle{plain}
 \newtheorem{theorem}{Theorem}%
 \newtheorem{proposition}[theorem]{Proposition}
 \newtheorem{lemma}[theorem]{Lemma}
 \newtheorem{corollary}[theorem]{Corollary}

 \theoremstyle{definition}

\AddToHook{env/theorem/begin}{\crefalias{theorem}{theorem}}
\AddToHook{env/proposition/begin}{\crefalias{theorem}{proposition}}
\AddToHook{env/lemma/begin}{\crefalias{theorem}{lemma}}
\AddToHook{env/corollary/begin}{\crefalias{theorem}{corollary}}
\AddToHook{env/definition/begin}{\crefalias{theorem}{definition}}
\AddToHook{env/remark/begin}{\crefalias{theorem}{remark}}

\def\eqref#1{equation~\ref{#1}}

\def\1{\bm{1}}

\DeclareMathAlphabet{\mathsfit}{\encodingdefault}{\sfdefault}{m}{sl}
\SetMathAlphabet{\mathsfit}{bold}{\encodingdefault}{\sfdefault}{bx}{n}

\newcommand{\R}{\mathbb{R}}

\title{Tight Transition Time Bounds for Separable Logistic Regression at the Edge of Stability}

\begin{document}

\maketitle
\suppressfloats[t]

\begin{abstract}
We study logistic regression on linearly separable data under gradient descent with a large constant stepsize $\eta$.
Such dynamics may exhibit a characteristic Edge of Stability  phenomenon, in which the loss initially oscillates before transitioning to a stable phase of monotone decrease.
Existing work provides a tight $\Theta(1)$ bound in dimension $d=2$ as $\eta \to \infty$ and conjectures a bound independent of $\eta$ in arbitrary dimensions $d\geq 2$.
In this paper, we disprove this conjecture by showing that, for every fixed sample size $n\geq 2$ and sufficiently small margin $\gamma$, the worst-case transition time is
$$\Theta\!\left((\log\eta)^{\min\{n-2,d-2\}}\right)$$
uniformly over $d\geq2$.
The key challenge in establishing a tight bound is that the sample contributing most strongly to the gradient can change repeatedly across iterations.
To address this issue, we control such changes by induction on dimension and sample size, and construct matching hard instances.

\end{abstract}

\section{Introduction}

Classical guarantees for gradient descent (GD) typically require small stepsizes: for a $\beta$-smooth objective, a stepsize $0<\eta<2/\beta$ guarantees monotone loss decrease~\citep{nesterov2018lectures}.
However, full-batch neural network training often operates in \emph{Edge of Stability} (EoS) regimes, where the largest Hessian eigenvalue hovers near $2/\eta$ and the loss oscillates over short time scales while decreasing over longer time scales~\citetext{\citealp{cohen2021gradient}; \citealp{cohen2025central}}.
Logistic regression on linearly separable data under GD with a large stepsize offers a tractable setting for studying such EoS regimes~\citetext{\citealp{wu2023implicit}; \citealp{wu2024large,crawshaw2026tight}}.

\begin{figure}[t]
\centering
\includegraphics[width=\linewidth]{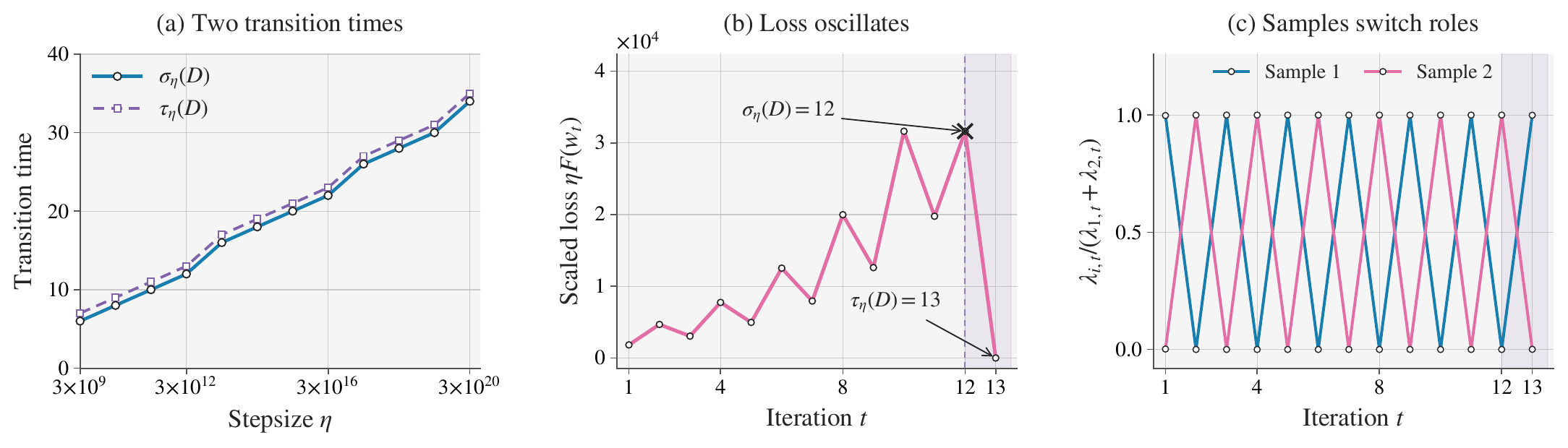}
\caption{\textbf{The transition time can grow logarithmically with the stepsize when $n=d=3$.}
(a) Both transition times $\sigma_\eta(D)$ and $\tau_{\eta}(D)$ grow approximately linearly with $\log\eta$, with a separate dataset for each stepsize and a common positive margin lower bound. (b,c) The loss continues to oscillate after all samples are correctly classified, as the two interacting samples alternate in driving the updates.
Shading marks the stable phase. See \Cref{sec:simulation-setup} for the details of the simulation.}
\label{fig:eos-oscillatory}
\end{figure}

In this regime, GD with a large constant stepsize can exhibit two phases: an initial EoS phase with oscillatory loss, followed by a stable phase of nonincreasing loss~\citep{wu2024large}.
To quantify when the transition {between these two phases} occurs, we define the \emph{exact EoS transition time} \(\sigma_\eta(D)\) as the earliest iterate after which the loss remains nonincreasing on dataset $D$ {with stepsize $\eta$}. Understanding its dependence on the stepsize $\eta$ is crucial for the optimization of GD within a finite time horizon: { A larger stepsize $\eta$ accelerates convergence in the stable phase where loss decreases monotonically with time. However, if increasing $\eta$ also extends the EoS phase, fewer iterations remain for stable descent within a fixed budget, potentially reducing the acceleration in the stable phase. Thus, sharp control of the transition time helps determine how large a constant stepsize can be used effectively within a fixed iteration budget.}

The sharp dependence of the transition time on the stepsize has remained open. \citet{wu2024large} show that GD reaches a sufficient condition for permanent monotonicity within $O(\eta)$ iterations, leaving open whether a {linear} dependence on the stepsize is necessary in the worst case. \citet{crawshaw2026tight} sharpen this bound in data dimension $d=2$: the first time the loss reaches a small threshold guaranteeing subsequent monotonicity is bounded independently of $\eta$ for sufficiently large $\eta$.
Based on numerical experiments, they conjecture that such a uniform bound independent of stepsize holds in arbitrary dimension.

In this paper, we disprove\footnote{\citet{crawshaw2026tight} study a threshold-based transition time, whereas we define the exact transition time. We show below that these two notions have the same worst-case asymptotic dependence on $\eta$.} this conjecture by establishing a tight worst-case dependence on $\eta$ for arbitrary $d$.
After absorbing the labels into the features, let $\mathcal D_{n,d}(\gamma)$ denote datasets of $n$ feature vectors $x_i\in\mathbb R^d$ with $\|x_i\|\leq 1$ for which there exists a unit separator with margin at least $\gamma$.
We consider GD from $w_0=0$ on the empirical logistic loss $F$.
For every fixed $n\geq2$, there exists $\gamma_n\in(0,1]$ such that, for every fixed $0<\gamma\leq\gamma_n$, Theorem~\ref{thm:main} gives
\[
\sup_{D\in\mathcal D_{n,d}(\gamma)}
\sigma_\eta(D)
=
\Theta\!\left(
(\log\eta)^{\min\{n-2,d-2\}}
\right)
\qquad\text{as }\eta\to\infty,
\]
uniformly over $d\geq2$. Both the constants and the stepsize threshold depend only on $n,\gamma$ and are independent of dimension $d$, while the worst-case dataset may depend on $\eta$.
Thus the worst-case transition time is bounded independently of $\eta$ when $n=2$ or $d=2$, but grows polylogarithmically once both $n,d\geq3$. Each increase in $\min\{n,d\}$ permits one additional logarithmic factor.
\Cref{fig:eos-oscillatory} illustrates the case $n=d=3$: the simulated transition times grow approximately linearly with $\log\eta$, and the loss can keep oscillating after every sample is correctly classified.
The construction and simulation setup are given in \Cref{sec:simulation-setup}.

The above main result has two corollaries.
\textbf{First, it disproves the conjecture of \citet{crawshaw2026tight}} that, for fixed sample size and margin, GD reaches the loss threshold $1/(8\eta)$ within a number of iterations bounded independently of $\eta$ in arbitrary dimension.
To state this corollary, let $k=\min\{n-2,d-2\}$ and define
\[
\tau_\eta(D)
=
\inf\left\{t\in\mathbb Z_{\geq0}:F(w_t)\leq\frac{1}{8\eta}\right\}.
\]
Under the same assumptions, Corollary~\ref{cor:threshold} gives
$\sup_{D\in\mathcal D_{n,d}(\gamma)}\tau_\eta(D)
=\Theta((\log\eta)^k)$, as $\eta \to \infty$, uniformly over $d\geq 2$.
This corollary establishes the same worst-case order for the threshold transition time $\tau_\eta(D)$ as our main theorem does for the exact EoS transition time $\sigma_\eta(D)$.
The corollary also indicates that {transition} times $\sigma_\eta(D)$ and $\tau_\eta(D)$ may differ on an individual dataset, but their worst-case orders agree. In summary, the stepsize independence established by \citet{crawshaw2026tight} is specific to the regime $d=2$ rather than a dimension-free phenomenon. 

\textbf{Second, this transition time bound yields a stronger guarantee for the optimization within a finite time horizon.}
We recall that the previous $O(\eta)$ transition bound, together with {the $\widetilde O(1/(\eta t))$ loss bound after $t$ further iterations in the stable phase}~\citep{wu2024large}
allows a stepsize $\eta=\Theta(T)$ and gives {final} loss $\widetilde O(T^{-2})$ after $T$ iterations.
In contrast, our polylogarithmic transition bound allows a much larger constant stepsize. For \(n,d\geq3\) and every fixed \(0<\gamma\leq1\), choosing $\eta=\exp(aT^{1/k})$ for a prescribed horizon \(T\), with \(a>0\) sufficiently small, ensures that \(\tau_\eta(D)\leq T/2\) for all sufficiently large \(T\). Recall that $k=\min\{n-2,d-2\}$, then applying the loss bound in the stable phase over the remaining iterations gives {the bound on the last-iterate loss $F(w_T)$ as }
\[
F(w_T)\leq\frac{C}{T}\exp\!\left(-aT^{1/k}\right),
\]
where $a,C>0$ and the required lower bound on $T$ depend only on $n,\gamma$ (Corollary~\ref{cor:optimization}). In contrast, \citet{wu2024large} obtain an $\widetilde O(T^{-2})$ bound on $F(w_T)$ in general dimension by choosing $\eta=\Theta(T)$.

\subsection{Technical Overview}
Proving a sharp transition bound requires going beyond two existing approaches.
\citet{wu2024large} obtain an $O(\eta)$ transition bound by combining progress along a separating direction with a global bound on iterate growth.
This establishes eventual stability but leaves the sharp stepsize dependence unresolved.
\citet{crawshaw2026tight} sharpen the analysis in $d=2$ by tracking oscillations in the one-dimensional subspace orthogonal to the max-margin direction.
Controlling these oscillations yields a stepsize-independent bound.
In higher dimensions, this orthogonal subspace has dimension $d-1>1$, so the same one-dimensional oscillation argument does not apply directly.

The key challenge in higher dimensions is that the samples contributing most strongly to the gradient can change repeatedly: a sample with a negligible current contribution can later become significant again, and this can happen repeatedly throughout the training trajectory.
Our analysis handles these changes through recursive projections that reduce the rank of the projected sample vectors.
Each recursive rank reduction contributes at most one logarithmic factor, while the lower bound constructs instances in which all such factors occur.

\textbf{Upper bound.}
The bound follows by controlling how long the dynamics can be driven by a given subset of samples and how often samples outside this subset can begin contributing significantly to the updates.
The \(O(\eta)\) analysis of \citet{wu2024large} relies on a global progress argument and does not exploit the finer structure of the oscillatory trajectory, while \citet{crawshaw2026tight} exploit this structure in \(d=2\), showing that oscillations force faster progress.
Our key insight is that \emph{after an appropriate projection, one can retain a nonempty subset of samples that remains separable while its rank decreases by at least one}, allowing us to bound the resulting dynamics recursively by induction.
{To bound the full trajectory before $\tau_\eta(D)$, we also control how often this reduction must be restarted.
We restart the reduction whenever a large update occurs or a sample outside the retained subset becomes significant again, and show that only $O(1)$ such restarts are needed.}

It therefore remains to bound the number of iterations between consecutive restarts. We call the consecutive iterations between restarts an \emph{outer block}. Within each outer block, we recursively reduce the rank of the projected sample vectors, that is, the dimension of the subspace they span.
At each recursive level, we partition the projected trajectory into intervals such that each completed interval increases the component along a separating direction by a fixed multiplicative factor. Since this component is globally bounded by $O(\eta)$, its multiplicative growth permits at most $O(\log\eta)$ intervals at each level. To bound the duration of each interval, we project again and retain a smaller subset of samples that remains separable and has lower rank. The induction then bounds the number of iterations in that interval. Summing these bounds over $O(\log\eta)$ intervals introduces at most one additional logarithmic factor.

The recursion ends when the projected sample vectors have rank one, where the duration is $O(1)$. For original data of rank $r\geq2$, the first projection has already reduced the rank to at most $r-1$, leaving at most $r-2$ recursive rank reductions.
Each contributes at most one logarithmic factor, and summing over the $O(1)$ outer blocks does not change the order.
Together with the direct bound for original rank one, this gives at most $\max\{r-2,0\}$ logarithmic factors.

\textbf{Lower bound.}
The key insight for the construction of hard instances is that \emph{a sample can contribute extremely little to the current update and still affect the transition time.}
When this contribution is extremely small, the updates are driven almost entirely by the remaining samples, while those updates continue to change the sample's margin.
Our construction makes the margin move only a constant fraction of the remaining distance before the sample becomes influential, allowing \(\Theta(\log\eta)\) repetitions.
Each additional interacting sample therefore contributes one additional factor of \(\log\eta\) to the total trajectory length.
This mechanism makes the transition time grow as a power of $\log\eta$, even for a fixed sample size and a fixed positive margin lower bound. 

More precisely, the exponent of $\log \eta$ is determined by the sample size and dimension. For $k\geq1$, producing $k$ logarithmic factors uses $k+1$ interacting samples in $k+1$ dimensions.
The recursive construction starts from a chosen initial state, which we realize after the first GD update from $w_0=0$ by adding one setup sample in an extra direction.
Thus, for $n,d\geq3$, we can take $k=\min\{n-2,d-2\}$.
However, a long trajectory alone does not establish a lower bound on the exact transition time, since the loss could already be decreasing monotonically. We therefore arrange a strict loss increase at time $\Theta((\log\eta)^k)$. This forces $\sigma_\eta(D)$ to occur later, giving the matching lower bound for the exact transition time.

\section{Related Work}
\label{sec:related}

\paragraph{Comparison with previous works on the transition to stable phase.}
The works most closely related to ours study large-stepsize GD on
separable logistic regression.
\citet{wu2024large} show that GD reaches a sufficiently stable regime within
\(O(\eta)\) iterations.
After the transition with additional $t$ steps, they obtain an
\(\widetilde O(1/(\eta t))\) loss guarantee, which leads to an
\(\widetilde O(T^{-2})\) rate when the constant stepsize is chosen on the order of the optimization horizon as \(\eta = \Theta(T)\). Their result demonstrates that instability can improve optimization, but
the \(O(\eta)\) transition bound leaves open whether the unstable phase
can grow polynomially with the stepsize.

\citet{crawshaw2026tight} substantially sharpen this picture when the data dimension $d = 2$.
They study the first time the loss reaches a sufficiently small threshold
that guarantees subsequent nonincreasing loss, and show that this time can
be bounded independently of \(\eta\) once the stepsize is sufficiently
large.
This raises the natural question of whether the same phenomenon persists
in higher dimensions.

We resolve this problem by giving the tight worst-case dependence on
\(\eta\) for arbitrary sample size and dimension.
We study the exact transition time, defined as the earliest time after which the loss remains nonincreasing, which more directly captures the dynamical transition than entry into a sufficient low-loss region.
Under the assumptions of Theorem~\ref{thm:main}, the two notions have the same worst-case asymptotic order.
The worst-case transition time is uniformly bounded when \(n=2\) or \(d=2\), but in
higher-dimensional instances it can grow as
\(\Theta((\log\eta)^{\min\{n-2,d-2\}})\).
Thus, the stepsize independence observed in two dimensions is a
low-dimensional phenomenon rather than a dimension-free property.

\paragraph{Edge of stability.}
\citet{cohen2021gradient} observed that full-batch neural network training often operates at EoS, where the largest Hessian eigenvalue stays near \(2/\eta\) and the loss oscillates while decreasing over longer time scales.
Other work studies the mechanisms sustaining these dynamics~\citep{lewkowycz2020large,arora2022understanding,damian2022self,cohen2025central} and their behavior in simplified models~\citep{chen2023beyond,zhu2023understanding,ahn2023learning,even2023sgd}.
Related analyses account for adaptive preconditioning and stochastic gradients~\citep{cohen2022adaptive,mulayoff2024exact,andreyev2024edge}, with recent empirical work examining EoS in language model pretraining~\citep{cai2026llm,meterez2026defense}.
These studies focus on training dynamics and local stability.
We instead study when GD on separable logistic regression enters a phase of permanent monotone loss decrease.

\paragraph{Separable logistic regression.}
On separable data, logistic loss has no finite minimizer, but GD with sufficiently small stepsizes converges in direction to the max-margin separator~\citep{soudry2018implicit}.
Related work studies convergence rates and loss tails~\citep{ji2019implicit,nacson2019convergence}, regularization paths~\citep{rosset2003margin,ji2020gradient}, and dual characterizations~\citep{ji2021characterizing}.
Stochastic gradient descent also drives the loss to zero with a sufficiently small fixed stepsize~\citep{nacson2019stochastic}, while finite iteration guarantees relate margins to generalization~\citep{shamir2021gradient}.
Under additional assumptions, \citet{wu2023implicit} extend convergence and implicit bias guarantees to arbitrary constant stepsizes despite possible loss oscillations.
For nonseparable data, GD can instead approach periodic trajectories~\citep{meng2024nonseparable,meng2025sphere}.
For separable data, \citet{tyurin2025perceptron} relates the large stepsize limit to a batch perceptron and studies how quickly GD finds a separator, which need not mark the end of the EoS phase.

\section{Main Results}
This section formalizes the exact EoS transition time and presents the main results.
Starting from the basic notation and necessary definitions in \Cref{sec:setup}, we first establish its tight dependence on the stepsize in \Cref{sec:mainresults}, and then derive two corollaries concerning the threshold transition time and finite-horizon optimization in \Cref{sec:corollaries}.

\subsection{Problem Setup}
\label{sec:setup}

\paragraph{Notation.}
We consider binary logistic regression with \(n\) samples in
\(\mathbb{R}^d\), where \(n,d\geq2\).
After absorbing the labels into the feature vectors, we write
\(D=(x_1,\ldots,x_n)\).
For \(0<\gamma\leq1\), let \(\mathcal D_{n,d}(\gamma)\) denote the class
of datasets for which there exists a unit vector \(w^\star\) satisfying
\[
\|x_i\|\leq1,
\qquad
\langle w^\star,x_i\rangle\geq\gamma,
\qquad i=1,\ldots,n.
\]
Repeated samples are allowed, and \(\gamma\) is a lower bound on the separator margin.
All logarithms are natural, and all norms and inner products are Euclidean.

\paragraph{Gradient descent.}
For \(D\in\mathcal D_{n,d}(\gamma)\), we consider the empirical logistic
loss
\[
F(w)
=
\frac1n\sum_{i=1}^n
\log\!\left(1+\exp(-\langle w,x_i\rangle)\right),
\]
and gradient descent with a constant stepsize \(\eta>0\),
\(w_{t+1}=w_t-\eta\nabla F(w_t)\), initialized at \(w_0=0\).

\paragraph{EoS transition time.}
For a fixed dataset \(D\), define the exact EoS transition time as
\[
\sigma_\eta(D)
:=
\inf\left\{
t\in\mathbb Z_{\geq0}:
F(w_{j+1})\leq F(w_j)
\text{ for every }j\geq t
\right\},
\]
with the convention \(\inf\varnothing=\infty\).
Thus, \(\sigma_\eta(D)\) is the earliest time after which the loss remains
nonincreasing, rather than the first iteration at which it decreases.
It equals zero if the loss is nonincreasing from initialization.

\subsection{Tight Bounds on the EoS Transition Time}
\label{sec:mainresults}

This section characterizes the dependence of the worst-case EoS transition time on the stepsize.

\begin{theorem}[Tight EoS transition time]
\label{thm:main}
For every integer \(n\geq2\), there exists a constant \(\gamma_n\in(0,1]\)
such that, for every fixed \(0<\gamma\leq\gamma_n\), 
\[
\sup_{D\in\mathcal D_{n,d}(\gamma)}
\sigma_\eta(D)
=
\Theta\!\left(
(\log\eta)^{\min\{n-2,d-2\}}
\right)
\qquad\text{as }\eta\to\infty,
\]
uniformly over all integers \(d\geq2\), where the implicit constants and the threshold on \(\eta\) depend only on \(n\) and \(\gamma\), independent of dimension $d$.
\end{theorem}

The theorem shows that the worst-case EoS transition time can depend on the stepsize.
While it remains bounded independently of \(\eta\) when \(n=2\) or \(d=2\), it grows with \(\eta\) once both \(n,d\geq3\).
This dependence is nevertheless mild: it is only polylogarithmic in the stepsize.

The exponent \(\min\{n-2,d-2\}\) describes how sample size and dimension control this stepsize dependence.
Each unit increase in \(\min\{n,d\}\) allows one additional logarithmic factor, while increasing the larger parameter alone leaves the exponent unchanged.
In particular, for fixed \(n\), the exponent saturates once \(d\geq n\).

It is worth noting that the characterization is uniform in dimension \(d\): the multiplicative constants and the threshold on \(\eta\) may depend only on \(n,\gamma\).
The upper bound holds for every fixed \(0<\gamma\leq1\), while the matching lower bound is established for \(\gamma\leq\gamma_n\).
The proof further sharpens the upper-bound exponent to \(\max\{r-2,0\}\), where \(r=\operatorname{rank}(x_1,\ldots,x_n)\). Therefore, $d=2$ suffices for a stepsize-independent bound.

\subsection{Corollaries}
\label{sec:corollaries}

We derive two corollaries of the transition bounds in Theorem~\ref{thm:main}.
Corollary~\ref{cor:threshold} connects the exact EoS transition time to the threshold-based notion studied in prior work, while Corollary~\ref{cor:optimization} translates the transition bound into an optimization guarantee.

\paragraph{Threshold transition time.}
Prior work measures the transition through the first entry into a low-loss region that guarantees nonincreasing loss thereafter~\citep{crawshaw2026tight}.
Following this convention, define
\[
\tau_\eta(D)
:=
\inf\left\{
t\in\mathbb Z_{\geq0}:
F(w_t)\leq\frac1{8\eta}
\right\},
\]
with the same convention \(\inf\emptyset=\infty\).
Corollary~\ref{cor:threshold} implies that the same worst-case order holds for this threshold time.

\begin{corollary}[Threshold transition time]
\label{cor:threshold}
With \(\gamma_n\) as in Theorem~\ref{thm:main}, for every fixed
\(n\geq2\) and \(0<\gamma\leq\gamma_n\),
\[
\sup_{D\in\mathcal D_{n,d}(\gamma)}
\tau_\eta(D)
=
\Theta\!\left(
(\log\eta)^{\min\{n-2,d-2\}}
\right)
\qquad\text{as }\eta\to\infty,
\]
uniformly over all integers \(d\geq2\).
\end{corollary}

\citet{crawshaw2026tight} showed that \(\tau_\eta(D)\) is bounded independently of \(\eta\) when \(d=2\), and conjectured the same bound in higher dimensions based on experiments.
Corollary~\ref{cor:threshold} disproves this conjecture: for fixed \(n,d\geq3\) and \(0<\gamma\leq\gamma_n\), the worst-case threshold time grows polylogarithmically with \(\eta\). The exact transition time \(\sigma_\eta(D)\) records when permanent monotonicity begins, while \(\tau_\eta(D)\) records entry into a sufficient low-loss region. These times need not coincide for a fixed dataset, even though their worst-case asymptotic orders agree.

\emph{Proof Sketch of Corollary~\ref{cor:threshold}.}
The stability estimate in \Cref{app:prelim} guarantees permanent monotonicity once \(F(w_t)\leq2/\eta\), so \(\sigma_\eta(D)\leq\tau_\eta(D)\).
\Cref{app:upper} proves
\(\tau_\eta(D)\leq C_{n,\gamma}(\log\eta)^{\min\{n-2,d-2\}}\)
for all sufficiently large \(\eta\), uniformly over \(D\in\mathcal D_{n,d}(\gamma)\) and \(d\geq2\).
Theorem~\ref{thm:main} supplies the matching lower bound through \(\sigma_\eta(D)\leq\tau_\eta(D)\), proving the corollary.

\paragraph{Optimization rate.}
The upper bound on \(\tau_\eta(D)\) provided in \Cref{app:upper} yields an optimization guarantee for every fixed \(0<\gamma\leq1\).
For \(n,d\geq3\), let \(k=\min\{n-2,d-2\}\geq1\).
Choosing \(\eta=\exp(aT^{1/k})\) with \(a>0\) sufficiently small ensures \(\tau_\eta(D)\leq T/2\) for large \(T\).
The stable-phase estimate \(F(w_{\tau_\eta(D)+s})=O_\gamma(1/(\eta s))\), \(s\in\mathbb Z_{\geq1}\), then gives the following last-iterate guarantee.

\begin{corollary}[Optimization]
\label{cor:optimization}
Fix \(n\geq3\) and \(0<\gamma\leq1\).
There exist constants \(a,C>0\) and an integer \(T_0\), depending only on
\(n\) and \(\gamma\), such that the following holds for every \(d\geq3\)
and every \(D\in\mathcal D_{n,d}(\gamma)\).
Let \(k=\min\{n-2,d-2\}\).
For every integer \(T\geq T_0\), gradient descent with the constant stepsize $\eta=\exp\!\left(aT^{1/k}\right)$
satisfies
\[
F(w_T)
\leq
\frac{C}{T}
\exp\!\left(-aT^{1/k}\right).
\]
\end{corollary}

The stepsize is chosen before the run as a function of the prescribed horizon \(T\) and remains constant throughout optimization.
For fixed \(n,d,\gamma\), the corollary guarantees loss at most \(\epsilon\) within
\(O((\log(1/\epsilon))^k)\) gradient updates as \(\epsilon\to 0\).
This improves the \(\widetilde O(T^{-2})\) guarantee obtained from the \(O(\eta)\) transition bound of \citet{wu2024large}: the polylogarithmic transient permits an exponentially larger constant stepsize.

For completeness, when \(k=0\), we have \(\tau_\eta(D)=O_{n,\gamma}(1)\) for sufficiently large \(\eta\).
The same analysis gives \(F(w_T)=O_\gamma(1/(\eta T))\) once \(T\) exceeds twice the uniform transition bound.

\section{Proof Sketch of Theorem~\ref{thm:main}}
\label{sec:proof-sketch}

This section introduces the main ideas behind Theorem~\ref{thm:main}.
The upper and lower bounds are driven by two complementary recursive
mechanisms. For the upper bound, each logarithmic factor comes from
partitioning a projected trajectory into \(O(\log\eta)\) phases before
reducing its rank. For the lower bound, a pair of coordinates generates
the first logarithmic time scale. Concatenating orbit segments and adding
gate coordinates produces higher powers of \(\log\eta\).
Throughout, we write
\[
k=\min\{n-2,d-2\},
\qquad B=\frac\eta n,
\qquad
\lambda_{i,t}=\frac{B}{1+\exp(\langle w_t,x_i\rangle)}.
\]
For sufficiently large \(\eta\), the upper bound proves
\(\sigma_\eta(D)\le\tau_\eta(D)\lesssim(\log\eta)^k\)
uniformly over \(D\in\mathcal D_{n,d}(\gamma)\).
The lower bound constructs a matching example when \(0<\gamma\le\gamma_n\).
The full proof is deferred to \Cref{app:upper} and \Cref{app:lower}.

\paragraph{Why the exponent is \(\min\{n-2,d-2\}\).}

Each factor of \(\log\eta\) comes from one additional recursive level,
which allows the trajectory from the previous level to repeat
logarithmically many times.  The basic two-coordinate dynamics already
produce a logarithmic trajectory starting from a fixed compact region in coefficient space;
one additional sample and direction are used to reach this region from
\(w_0=0\).  Thus the first logarithmic factor appears when \(n,d\ge3\).
Thereafter, each additional factor requires one more interacting sample
and one more independent direction.  
Thus, the lower-bound construction realizes \(k=\min\{n-2,d-2\}\) logarithmic levels. The upper-bound rank recursion shows that no additional levels are possible, yielding the rate $\Theta\!\left((\log\eta)^{\min\{n-2,d-2\}}\right).$

\subsection{Upper Bound}

The main obstacle is that the loss need not decrease before the transition, so it cannot serve as a global potential. Our proof instead controls the trajectory through the coefficients \(\lambda_{i,t}\) in three steps.
First, we reduce the problem to reaching a region of permanently monotone loss.
Second, we partition the trajectory before \(\tau_\eta(D)\) into outer blocks, restarting whenever a large coefficient appears or a previously negligible sample becomes significant again, and use a global coefficient budget to show that only \(O(1)\) such blocks can occur.
Finally, within each outer block, we retain a subset of samples that remains separable after projection and whose projected rank is at least one smaller. We then bound the length of the resulting lower-rank process recursively.

\paragraph{Step~1. Reduce permanent monotonicity to reaching a small loss.} Lemma~\ref{lem:sketch-stability} shows that once the loss is sufficiently small, it remains nonincreasing at every subsequent iteration.
\begin{lemma}[\citet{wu2024large}]
\label{lem:sketch-stability}
If \(F(w_t)\le2/\eta\), then \(F(w_{j+1})\le F(w_j)\) for every \(j\ge t\).
\end{lemma}

Logistic loss satisfies \(\nabla^2F(w)\preceq F(w)I\), and hence
\(\nabla^2\log F(w)\preceq I\).
Expanding \(\log F\) to second order along one gradient step gives $\log F(w_{t+1})-\log F(w_t)
\le-\frac{\eta\|\nabla F(w_t)\|^2}{F(w_t)}
\left(1-\frac{\eta F(w_t)}2\right).$
Thus the loss is nonincreasing and stays at most \(2/\eta\), so the argument repeats.
Since \(\tau_\eta\) uses the smaller threshold \(1/(8\eta)\), this proves
\(\sigma_\eta(D)\le\tau_\eta(D)\).
\Cref{app:prelim} gives the details.

\paragraph{Step~2. Bounding the number of outer blocks.}
Our goal is to show that the trajectory before \(\tau_\eta\) can be partitioned into only \(O_{n,\gamma}(1)\) outer blocks.
A new block is needed only when an exceptional event occurs: either some coefficient becomes large, or a previously negligible sample becomes significant again.
Let \(c_t=\sum_i\lambda_{i,t}\) denote the coefficient mass at time \(t\). Lemma~\ref{lem:sketch-mass} gives a global bound on the cumulative coefficient mass, which limits how often a block can be restarted.

\begin{lemma}[Global coefficient mass]
\label{lem:sketch-mass}
For all sufficiently large \(\eta\), there exists a universal constant
\(c_0>0\) such that $c_t \geq c_0$ for all $t<\tau_\eta$. Moreover, it holds that $\sum_{t<\tau_\eta} c_t
\leq
\frac{2\eta}{\gamma^2} = O(\eta).
$
\end{lemma}

Lemma~\ref{lem:sketch-mass} provides the global budget needed to control the number of outer blocks. 
Every restart is associated with one of two events. 
The first is a large coefficient, \(\lambda_{i,t}\ge B/4\), which gives $c_t\ge B/4=\Omega(\eta)$ in a single update. 
The second is that a previously negligible sample becomes significant again.
The block construction below ensures that each such return also requires \(\Omega(\eta)\) cumulative coefficient mass.
Since Lemma~\ref{lem:sketch-mass} gives only $O(\eta)$ cumulative coefficient mass before  \(\tau_\eta\), only $O(1)$ restarts can occur, and hence there are only $O(1)$ outer blocks.

To prove Lemma~\ref{lem:sketch-mass}, track \(\|w_t-(2\eta/\gamma)w^\star\|^2\).
Before \(\tau_\eta\), this nonnegative quantity decreases by at least
\(2\eta c_t\) in each update. Its initial value is \(4\eta^2/\gamma^2\),
so summing these decreases over any finite prefix gives the \(O(\eta)\) coefficient budget in Lemma~\ref{lem:sketch-mass}.
The same argument also yields $\|w_t\|\le 4\eta/\gamma$ for $t \leq \tau_\eta$.
Moreover, the logistic formula gives \(c_t\ge c_0\) for 
\(t < \tau_\eta\), so the finite budget ensures that \(\tau_\eta<\infty\).
\Cref{sub:upperoriginal} supplies the details.

\paragraph{Step~3. Recursive control within each outer block.}
We bound the length of a single outer block by reducing it to a lower-rank process. 
Let \(r=\operatorname{rank}(x_1,\ldots,x_n)\).
The case \(r=1\) is immediate, so assume \(r\ge2\). 
At a block start \(s\ge1\), consider $u = \frac{w_s}{\|w_s\|}$, projector $P=I-uu^\top$ and retain the samples satisfying $\langle Pw^\star,Px_i\rangle\ge\frac{\gamma}{2}$.
The retained samples remain separable after projection, while their projected rank is at most \(r-1\). 
Thus, a single projection already reduces the rank by at least one. 
The remaining samples stay negligible throughout the block and contribute only a small additive error to the projected dynamics.

The projection removes the component along \(u\), but this component still contributes to the retained margins. Since all sample margins are positive at the block start, its contribution is nonnegative and nondecreasing throughout the block. This leads to the following projected process.

\begin{lemma}[Projected dynamics]
\label{lem:sketch-recursion}
Let \(y_1,\ldots,y_m\) be vectors of norm at most one and rank \(r'\), separated by a unit vector with margin \(\mu>0\). Consider
$$M_{i,t}=b_i+h_{i,t}+\langle z_t,y_i\rangle,
\qquad \lambda_{i,t}=\frac{B}{1+\exp(M_{i,t})},
\qquad z_{t+1}=z_t+\sum_i\lambda_{i,t}y_i+e_t,$$
where $z_0=0$, \(b_i\ge\log 3\), \(h_{i,0}=0\), and each \(h_{i,t}\) is nonnegative and nondecreasing. Assume that, for \(0\le t<T\), 
$
\lambda_{i,t}<B/4,
\sum_i\lambda_{i,t}\ge c>0,
\|z_t\|=O(B)$, and $
\|e_t\|=O(B^{-6}).
$
Then, for sufficiently large \(B\), it holds that
$
T\le C(1+\log B)^{r'-1},
$
where \(C\) and the threshold on \(B\) depend only on
\(m,\mu,c\) and the constants in the two norm bounds.
\end{lemma}

The projected dynamics within each original block satisfy the assumptions of Lemma~\ref{lem:sketch-recursion}: the removed direction contributes a nondecreasing offset, while the discarded samples contribute only the allowed small error. Since the projected rank is at most \(r-1\), the lemma gives a block length of $O_{n,\gamma}\!\left((1+\log B)^{r-2}\right).$

The lemma itself is proved recursively in \Cref{sub:upperaux}. When \(r'=1\), the coefficient mass falls below \(c\) within \(O(1)\) updates. For \(r'\ge2\), after at most \(O(\log B)\) initial updates, the trajectory is partitioned into \(O(\log B)\) intervals. Within each interval, another projection produces the same process class with fewer samples and rank at most \(r'-1\). This leads to $ T_{r'}(B)
\lesssim
(1+\log B)\,T_{r'-1}(B),
$
and therefore
$
T_{r'}(B)
=
O\!\left((1+\log B)^{r'-1}\right).
$
Together with the \(O_{n,\gamma}(1)\) outer blocks and exceptional updates from Step~2,
$
\tau_\eta(D)
\le
C_{n,\gamma}(\log\eta)^{\max\{r-2,0\}}.
$
Since \(r\le\min\{n,d\}\), the desired upper bound follows.

\subsection{Lower Bound}

Let \(k=\min\{n-2,d-2\}\).
The proof has three steps.
First, we construct a two-coordinate trajectory of logarithmic length.
Second, we add coordinates recursively, with each new coordinate introducing one additional logarithmic factor, yielding a trajectory of length \(\Theta((\log B)^k)\).
Third, we show that the loss still increases near the end of this trajectory and realize the construction by a separable dataset initialized at \(w_0=0\).
When \(k=0\), a separate example gives a strict loss increase at the first update, so \(\sigma_\eta\ge1\); see \Cref{sub:lowercomplete}.

\paragraph{Step~1. Constructing the base logarithmic trajectory.}
We begin with two sample vectors whose pairwise inner products are specified by the Gram matrix $K_1$, defined by $(K_1)_{ij}=\langle x_i,x_j\rangle$.
We choose
\[
K_1=
\begin{pmatrix}
1&-a\\
-a&1
\end{pmatrix},
\qquad a=\frac45.
\]
For any Gram matrix \(K\), define
\(p_i=B\exp(-\langle w,x_i\rangle)\) and
\(h_B(x)=x/(1+x/B)\). Since the actual update coefficient is \(\lambda_i=h_B(p_i)\), the induced dynamics are
\[
p^+=p\odot\exp(-Kh_B(p)),
\]
where the product, exponential, and \(h_B\) act coordinatewise.
For \(K_1\), when one \(p_i\) is large and the other is small, the negative off-diagonal entries make the next update decrease the large one and increase the small one. This alternating behavior is the basic mechanism used in the construction.

\begin{lemma}[Base logarithmic trajectory]
\label{lem:sketch-base}
For all sufficiently large \(X\) satisfying \(X^2\le B\), there exists
a trajectory of the \(K_1\) dynamics with length \(\Theta(\log X)\)
whose initial state lies in a fixed compact subset of
\((0,\infty)^2\) and whose final state has one coordinate equal to \(X\).
\end{lemma}

Taking \(X=B^{1/4}\), Lemma~\ref{lem:sketch-base} gives a trajectory of
length \(\Theta(\log B)\) starting from a fixed bounded region.  This is
the first logarithmic time scale in the lower-bound construction.
A long trajectory alone does not yet imply a lower bound on the
transition time; in Step~3, we will show that a strict loss increase occurs
near the end of the constructed trajectory.

We construct the trajectory backward.  Starting from a state with one
large coordinate and one suitably chosen exponentially small coordinate,
the exact inverse update exchanges their roles: the previously small
coordinate becomes large, while the large value decreases from \(X\) to
\(aX+O(1)\).  
The resulting state has the same structure, so the same
backward step can be repeated.  Since \(0<a<1\), the large coordinate
decreases geometrically until it reaches a fixed constant range, which
takes \(\Theta(\log X)\) steps.  Reversing the sequence gives the
trajectory in Lemma~\ref{lem:sketch-base}.  \Cref{sub:lowerorbit},
Step~1, gives the exact inverse calculation.

\paragraph{Step~2. Recursively adding logarithmic scales.}
The key idea is to make the previous-level trajectory repeat
logarithmically many times.  Each additional coordinate enables such a
repetition, introducing one additional logarithmic factor in the
trajectory length. This leads to Lemma~\ref{lem:sketch-hierarchy}.

\begin{lemma}[Recursive logarithmic trajectory]
\label{lem:sketch-hierarchy}
For each fixed \(r\ge2\), there exists a Gram matrix
\(K_r\) of size \(r+1\) such that, whenever \(E\) is sufficiently large
and \(E\le B^{1/4}\), the \(K_r\) dynamics admit a trajectory segment of
length $\Theta\!\left((\log E)^{r-1}\right)$,
whose final state has last coordinate \(e^{-E}\) and whose initial gate coordinate is \(e^{-D}\), where $\frac{E}{\Lambda_r}\le D\le \frac{E}{2}$ for some constant \(\Lambda_r\ge4\). The remaining coordinates at both endpoints lie in fixed compact sets independent of \(E\).
\end{lemma}

The construction is uniform over the bounded endpoint profiles that arise recursively: each segment starts from another admissible bounded profile, so the segments can be concatenated.
We prove Lemma~\ref{lem:sketch-hierarchy} by induction.
For the base case \(r=2\), we add a third coordinate to the trajectory from Step~1.  
This coordinate remains exponentially small while the two-coordinate dynamics evolve, and this gives a trajectory of length \(\Theta(\log E)\).

For the induction step, suppose the level-\(r\) construction has length \(\Theta((\log E)^{r-1})\).
We add one more coordinate and couple it weakly to the last coordinate of the existing system.
The final update sends the new coordinate from a value of order \(E\) to \(e^{-E}\).
Through the negative coupling, solving this update backward places the last coordinate of the old system at \(e^{-\Theta(E)}\).
We then concatenate level-\(r\) segments backward, with the exponent in this coordinate decreasing by a fixed factor after each segment.
This gives \(\Theta(\log E)\) segments in total, whose lengths sum to \(\Theta((\log E)^r)\).
Along these concatenated segments, the new coordinate remains
exponentially small, so its perturbation to the previous-level dynamics can be corrected backward to obtain an exact trajectory.

Finally, for \(k\ge2\), we start with \(E=B^{1/4}\) and concatenate
level-\(k\) segments backward until the exponent reaches a fixed
constant range.  This contributes one final logarithmic factor and gives
\[
T=\Theta\!\left((\log B)^k\right).
\]
The resulting trajectory starts in a fixed compact subset of the
positive orthant and has total coefficient mass \(O(B^{1/4})\).
For \(k=1\), Step~1 gives the same conclusion with
\(T=\Theta(\log B)\).
The full construction is provided in \Cref{sub:lowerorbit}.

\paragraph{Step~3. Forcing a late loss increase.}
We next show that the loss still increases near the end of the trajectory, since a long trajectory alone does not imply a late transition.
For the auxiliary dynamics, define the scaled loss $L_B(p)=B\sum_i\log(1+p_i/B).$
When the auxiliary coordinates correspond to samples in the original
problem, \(L_B\) equals \(\eta\) times their contribution to \(F\).

\begin{proposition}[Late loss increase]
\label{prop:sketch-late}
For each fixed \(k\ge1\) and sufficiently large \(B\), the constructed
auxiliary orbit has \(L_B(p_{t+1})>L_B(p_t)\) at some
\(t=\Theta((\log B)^k)\).
\end{proposition}

For \(k\ge2\), consider the last level-\(k\) segment constructed in
Step~2.  The full trajectory has length
\(\Theta((\log B)^k)\), whereas this final segment has length only
\(O((\log B)^{k-1})\).  Hence this segment begins at time
\(\Theta((\log B)^k)\).
At the beginning of the segment, the scaled loss is \(O(1)\). 
During the segment, the final gate produces a coefficient of order \(B^{1/4}\), and hence its coordinate reaches order \(B^{1/4}\) at some later time. 
The scaled loss is therefore \(\Omega(B^{1/4})\) at that time. Thus at least one update within the segment strictly increases the scaled loss.
Since the segment itself begins at time
\(\Theta((\log B)^k)\), this increase also occurs at
\(t=\Theta((\log B)^k)\).
For \(k=1\), the final update of the two-coordinate trajectory gives the
strict increase directly.

It remains to realize the auxiliary trajectory from \(w_0=0\).
We add setup samples in one extra direction so that the first update
reaches the required auxiliary initial state exactly.
These samples then acquire margins of order \(B\) and remain negligible
throughout the trajectory, since the total subsequent coefficient mass
is only \(O(B^{1/4})\).
Hence the true gradient-descent trajectory stays close to the auxiliary
one, so the late loss increase persists and
$\sigma_\eta(D)\gtrsim(\log B)^k.$
The construction also preserves bounded sample norms and a positive
separation margin independent of \(B\).

\section{Conclusion}

We established tight worst-case bounds for the Edge-of-Stability transition time of constant-stepsize gradient descent on separable logistic regression, showing that it scales as \(\Theta((\log \eta)^{\min\{n-2,d-2\}})\) for sufficiently small fixed margin. Thus, stepsize independence holds when \(n=2\) or \(d=2\), while in higher-dimensional settings the EoS phase can grow polylogarithmically with the stepsize. The same order holds for the threshold transition time, yielding substantially faster finite-horizon optimization with large constant stepsizes.

Our analysis further shows that each logarithmic factor corresponds to an additional level of interaction in the dynamics. The upper bound limits these levels through recursive rank reduction, while the lower bound constructs instances that realize all of them. Together, these results give a sharp characterization of how sample size and data dimension govern the duration of large-stepsize instability.
More broadly, it remains to understand whether analogous hierarchical mechanisms govern large-stepsize transitions beyond separable logistic regression.

\par
\section{Acknowledgement}
We thank Kaifeng Lyu and Huanran Chen for insightful discussions.

\newpage

\tableofcontents

\newpage

\clearpage
\appendix
\begin{huge}
\begin{center}
    Appendix
\end{center}
\end{huge}

\section{Illustration of Theory}
\label{sec:simulation-setup}

\begin{figure}[!ht]
\centering
\includegraphics[width=\linewidth]{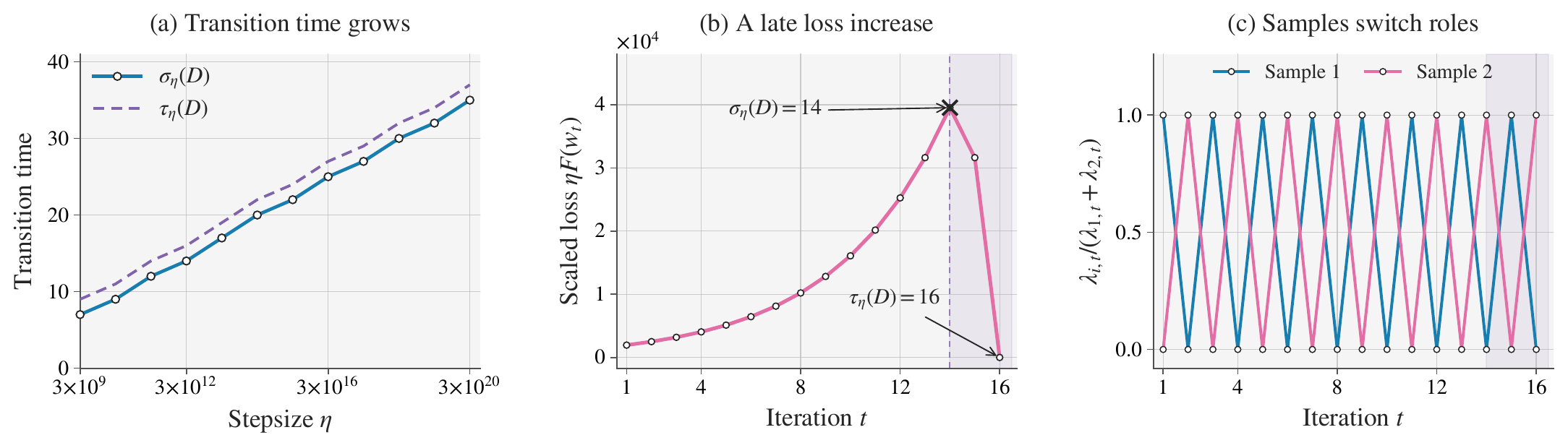}
\caption{\textbf{The total loss can keep increasing while the dominant sample switches.}
For the symmetric construction with $n=d=3$,
(a) both transition times grow approximately linearly with $\log\eta$, with a separate dataset for each stepsize.
(b,c) Even after all samples are correctly classified, the loss keeps increasing as the two interacting samples alternate in driving the updates.
Panel (c) shows their relative update coefficients.
Shading marks the stable phase.}
\label{fig:eos-symmetric}
\end{figure}

We simulate GD from $w_0=0$ with $n=d=3$, the smallest sample size and dimension for which the transition bound can grow with $\eta$.
\Cref{fig:eos-symmetric} uses the symmetric pair construction in \Cref{sub:lowerorbit,sub:lowerrealization}, while \Cref{fig:eos-oscillatory} uses a variant with unequal sample norms.
Both figures report forward GD on three separable samples.

\paragraph{Two separable constructions.}
We fix $\rho=0.1$ and write $B=\eta/3$ and $\widehat B=\rho^2B$, following \Cref{sub:lowerrealization}.
The first two samples are $x_i=(v_i,0)\in\mathbb R^3$, where $v_i\in\mathbb R^2$.
For the oscillating loss in \Cref{fig:eos-oscillatory}, we use
\[
v_1=\rho(1,0),\qquad v_2=\rho(-0.4,0.3),\qquad
K=\begin{pmatrix}1&-0.4\\-0.4&0.25\end{pmatrix}.
\]
For \Cref{fig:eos-symmetric}, we use
\[
v_1=\rho(\sqrt{0.1},\sqrt{0.9}),\qquad
v_2=\rho(\sqrt{0.1},-\sqrt{0.9}),\qquad
K=\begin{pmatrix}1&-0.8\\-0.8&1\end{pmatrix}.
\]
In each case, $\langle v_i,v_j\rangle=\rho^2K_{ij}$.
The negative inner product means that an update along one sample reduces the other sample's margin.
The pair geometry stays fixed as $\eta$ varies.
Only the third sample changes to realize the initial state constructed below.

\paragraph{Constructing the auxiliary trajectory backward.}
We use the logistic coefficient map from \Cref{sub:lowerdynamics} at scale $\widehat B$.
Let $m_i$ denote the current margin of sample $i$ and set $\widehat p_i=\widehat B e^{-m_i}=\rho^2p_i$, where $p_i$ is defined in \Cref{sub:lowerdynamics}.
With $h_{\widehat B}(z)=z/(1+z/\widehat B)$, one update of the isolated pair is
\[
\widehat p^{+}=\widehat p\odot\exp\bigl(-K h_{\widehat B}(\widehat p)\bigr),
\]
where both functions act coordinatewise.
Set $X=\widehat B^{1/4}$ and $c=-K_{12}>0$.
At the terminal state, choose the dominant index $i=2$ for \Cref{fig:eos-oscillatory} and $i=1$ for \Cref{fig:eos-symmetric}, and let $j$ be the other index.
The terminal coordinates are
\[
\widehat p_i=X,\qquad
\widehat p_j=\frac{K_{ii}X}{c}\exp\bigl(-c h_{\widehat B}(X)\bigr).
\]
We repeatedly invert the logistic map until $\max_i\widehat p_i\leq20$.
At each backward step, let $i$ be the currently dominant index and $j$ the other index.

We solve for predecessor coordinates $u<v$ at indices $i$ and $j$:
\[
\begin{aligned}
\log u-K_{ii}h_{\widehat B}(u)+c h_{\widehat B}(v)&=\log\widehat p_i,\\
\log v-K_{jj}h_{\widehat B}(v)+c h_{\widehat B}(u)&=\log\widehat p_j.
\end{aligned}
\]
Newton iteration starts from $v=c\widehat p_i/K_{jj}$ and uses the small positive solution for $u$.
Reversing the resulting list gives an auxiliary trajectory with initial state $\widehat p_0$.
This procedure inverts the logistic map itself.

\paragraph{Realizing the initial state from zero.}
Set $\mu_i=\log(\widehat B/\widehat p_{i,0})$ and choose $u_B\in\mathbb R^2$ satisfying
\[
\langle v_i,u_B\rangle=\frac{2\mu_i}{B},\qquad i=1,2.
\]
We then add the setup sample
\[
x_3=(u_B-v_1-v_2,\,1/2).
\]
At $w_0=0$, all three update coefficients equal $B/2$, so the first GD update is
\[
w_1=\left(\frac B2u_B,\frac B4\right),\qquad
\langle w_1,x_i\rangle=\mu_i\quad(i=1,2).
\]
Thus, in exact arithmetic, the first update reaches the required initial margins exactly.
The setup sample then has a large positive margin and an exponentially small update coefficient.
It is included in every subsequent GD update, but the later dynamics are driven almost entirely by the first two samples.
This is why panel (c) shows their relative coefficients $\lambda_{i,t}/(\lambda_{1,t}+\lambda_{2,t})$, with $\lambda_{i,t}=B/(1+e^{\langle w_t,x_i\rangle})$.
The ratios describe scalar coefficients within this pair, not fractions of the full gradient vector.

For every simulated dataset, we verify $\|x_i\|\leq1$ and a common margin lower bound $\gamma=1/(40\sqrt{10})$.
The fixed unit separators are
\[
w^\star=\left(\frac{3}{2\sqrt{205}},\frac{7}{\sqrt{205}},\frac{\sqrt3}{2}\right)
\quad\text{for \Cref{fig:eos-oscillatory}},\qquad
w^\star=\left(\frac12,0,\frac{\sqrt3}{2}\right)
\quad\text{for \Cref{fig:eos-symmetric}}.
\]
The datasets may therefore vary with $\eta$ while $n$, $d$, and the positive margin lower bound remain fixed.

\paragraph{Measuring the transition times.}
We test $\eta=3\times10^9,3\times10^{10},\ldots,3\times10^{20}$.
Panels (b,c) use $\eta=3\times10^{12}$ and display every iterate from $t=1$ to $\tau_\eta(D)$.
We omit $t=0$ from these panels because $\eta F(w_0)=\eta\log2$ is much larger than the subsequent losses.
Every sample is correctly classified throughout the displayed trajectories.
We run forward until the first iterate with $\eta F(w_t)\leq1/8$, which gives $\tau_\eta(D)$.
By Lemma~\ref{lem:sketch-stability}, no later loss increase is possible.
We obtain $\sigma_\eta(D)$ as one plus the last index $t$ with $F(w_{t+1})>F(w_t)$, or zero if no such index exists.
The shaded regions therefore begin at $\sigma_\eta(D)$, not at $\tau_\eta(D)$.

\clearpage
\section{Additional Related Works}
\paragraph{Acceleration of gradient methods.}
Suitable stepsize schedules or momentum can improve margin convergence on separable data~\citep{nacson2019convergence,ji2021characterizing,ji2021fast}.
Adaptive or increasing stepsizes also accelerate logistic loss convergence~\citep{axiotis2023gradient,kale2026stretched}, including guarantees for averaged iterates~\citep{zhang2025gradient}.
For smooth convex objectives, schedules with long steps improve convergence~\citep{grimmer2024provably,altschuler2025silver}, with extensions providing guarantees at any stopping time~\citep{zhang2025anytime}.

\paragraph{Large stepsizes in related settings.}
The optimization benefits of large stepsizes extend to local GD with heterogeneous objectives~\citep{crawshaw2025constant} and regularized logistic regression~\citep{wu2025large}.
Related work gives transition guarantees for two layer networks under activation and separability assumptions~\citep{cai2024large}, and convergence with arbitrary constant stepsizes for a broader class of classification losses on separable data~\citep{bao2025any}.
In our setting, faster optimization follows from a tight bound on the EoS transition time with a fixed stepsize.

\providecommand{\ip}[2]{\langle #1,#2\rangle}
\providecommand{\norm}[1]{\lVert #1\rVert}
\providecommand{\spn}{\operatorname{span}}
\providecommand{\diag}{\operatorname{diag}}
\providecommand{\interior}{\operatorname{int}}
\providecommand{\e}{\mathrm e}
\newcommand{\proofstep}[1]{\par\medskip\noindent\textbf{#1}\par\smallskip}
\renewcommand{\eqref}[1]{\textup{(\ref{#1})}}

\section{Preliminaries and the monotone region}
\label{app:prelim}

Throughout the appendix, we use the data class, initialization, and
transition time $\sigma_\eta(D)$ from \Cref{sec:setup}.
The threshold time $\tau_\eta(D)$ is defined in
\Cref{sec:corollaries} by $F(w_t)\le1/(8\eta)$.
We write $B=\eta/n$, as in \Cref{sec:proof-sketch}.
In particular,
\[
F(w)=\frac1n\sum_{i=1}^n\ell(\ip{w}{x_i}),
\qquad
\ell(s)=\log(1+\exp(-s)),
\]
and $\norm{x_i}\le1$ for every sample.
All logarithms are natural. Vector norms are Euclidean, and matrix norms
are the induced operator norms. For symmetric matrices, $A\preceq C$
means $z^\top Az\le z^\top Cz$ for every vector $z$.
We omit $D$ from a transition time when the dataset is fixed.

\subsection{Proof of Lemma~\ref{lem:sketch-stability}}

\begin{proof}
Let
\[
q(s)=\frac1{1+\exp(s)}.
\]
Then $\ell'(s)=-q(s)$ and $\ell''(s)=q(s)(1-q(s))$.
Since
\[
q(s)\le \log(1+\exp(-s))=\ell(s),
\]
where we used $\log(1+x)\ge x/(1+x)$ for $x\ge0$, we obtain
\[
\ell''(s)\le \ell(s).
\]
Consequently,
\begin{align*}
\nabla^2F(w)
&=\frac1n\sum_{i=1}^n
\ell''(\ip{w}{x_i})x_ix_i^\top\\
&\preceq
\frac1n\sum_{i=1}^n
\ell(\ip{w}{x_i})I
=F(w)I.
\end{align*}
Therefore
\[
\nabla^2\log F(w)
=
\frac{\nabla^2F(w)}{F(w)}
-
\frac{\nabla F(w)\nabla F(w)^\top}{F(w)^2}
\preceq I.
\]
For any displacement $v$, apply the fundamental theorem of calculus twice
to $g(u)=\log F(w+uv)$ on $[0,1]$. The Hessian bound gives
\[
g(1)-g(0)
=g'(0)+\int_0^1(1-u)g''(u)\,\mathrm du
\le\frac{\ip{\nabla F(w)}{v}}{F(w)}+\frac12\norm v^2.
\]
Taking $v=-\eta\nabla F(w)$ and $w^+=w+v$ yields
\begin{align}
\log F(w^+)-\log F(w)
&\le
-\eta\frac{\norm{\nabla F(w)}^2}{F(w)}
+\frac{\eta^2}{2}\norm{\nabla F(w)}^2\notag\\
&=
-\eta\frac{\norm{\nabla F(w)}^2}{F(w)}
\left(1-\frac{\eta F(w)}2\right).
\label{eq:loglossdescent}
\end{align}
If $F(w)\le2/\eta$, the right hand side is nonpositive. Hence
$F(w^+)\le F(w)\le2/\eta$.  The sublevel set is therefore forward
invariant and the loss is nonincreasing at every subsequent iterate.
Since the threshold defining $\tau_\eta$ is $1/(8\eta)<2/\eta$, we also
have $\sigma_\eta(D)\le\tau_\eta(D)$.
\end{proof}

\section{Upper bound}
\label{app:upper}

We prove that, for each $n\ge2$ and $0<\gamma\le1$, there are constants
$C_{n,\gamma}>0$ and $\eta_{n,\gamma}$ such that
\begin{equation}\label{eq:appendixupper}
\tau_\eta(D)
\le C_{n,\gamma}(1+\log\eta)^{\min\{n-2,d-2\}}
\end{equation}
for every $d\ge2$, $D\in\mathcal D_{n,d}(\gamma)$, and
$\eta\ge\eta_{n,\gamma}$. Lemma~\ref{lem:sketch-stability} then gives
the upper bound in Theorem~\ref{thm:main}.

\subsection{An auxiliary interval estimate}\label{sub:upperaux}
Fix an integer $m\ge1$ and constants
\[
  0<\mu\le1,
  \qquad c>0,
  \qquad K>0,
  \qquad E\ge0.
\]
Let $y_1,\ldots,y_m$ be vectors in a finite dimensional Euclidean space with
\[
  \norm{y_i}\le1,
  \qquad \ip{v^\star}{y_i}\ge\mu,
  \qquad \norm{v^\star}=1.
\]
Within this subsection, write
\[
  r=\dim\spn\{y_1,\ldots,y_m\}.
\]
Thus $r$ is the rank of the samples in the auxiliary process.

Let $T$ be a nonnegative integer and $B>0$.
For $0\le t\le T$, define margins and coefficients by
\begin{equation}\label{eq:auxmargin}
  M_{i,t}=b_i+h_{i,t}+\ip{z_t}{y_i},
  \qquad
  \lambda_{i,t}=\frac{B}{1+\exp(M_{i,t})}.
\end{equation}
The numbers $b_i$ are fixed, and the numbers $h_{i,t}$ satisfy
\begin{equation}\label{eq:offsetassumptions}
  b_i\ge\log3,
  \qquad h_{i,0}=0,
  \qquad 0\le h_{i,t}\le h_{i,t+1}
  \qquad(0\le t<T).
\end{equation}
Assume that $z_0=0$ and
\begin{equation}\label{eq:auxupdate}
  z_{t+1}=z_t+\sum_{i=1}^m\lambda_{i,t}y_i+e_t
  \qquad(0\le t<T),
\end{equation}
with the bounds
\begin{equation}\label{eq:auxbounds}
  \lambda_{i,t}<\frac B4,
  \qquad \sum_{i=1}^m\lambda_{i,t}\ge c,
  \qquad \norm{z_t}\le KB,
  \qquad \norm{e_t}\le EB^{-6}
  \qquad(0\le t<T).
\end{equation}
The vectors $e_t$ are additive errors. No condition is imposed on how the
offsets or errors are chosen beyond the displayed assumptions.
The value of $B$ does not change when some samples are removed.

The margins and offsets are defined through time $T$, but the bounds in
\eqref{eq:auxbounds} are required only before $T$. In particular, neither a
coefficient bound nor a norm bound is assumed at the terminal state $z_T$.
This distinction allows the last update of an interval to be included even
when that update ends the interval.

We claim that there are constants $C>0$ and $B_0<\infty$, depending only
on $m,\mu,c,K,E$, such that every process just described with $B\ge B_0$ satisfies
\begin{equation}\label{up:auxbound}
 T\le C(1+\log B)^{r-1}.
\end{equation}
Because $r\le m$, the exponent is also $\min\{m-1,r-1\}$.
The constants are uniform over the sample vectors and their ambient dimension.
We now prove this claim in full.

We use strong induction on $m$. This means that, when proving the statement
for $m$ samples, we may apply it to every smaller sample count and every
choice of its scalar parameters. The case $r=1$ will be proved for every $m$.
It includes $m=1$ and therefore provides the base of the induction.

During the proof of this auxiliary estimate, a constant denoted by $C$ may
increase from one occurrence to the next. It depends only on $m,\mu,c,K,E$. Every lower threshold imposed on
$B$ will have the same permitted dependence. We may assume $B\ge\e$ and
$T\ge1$, since $T=0$ is immediate.

\proofstep{Step 1. Restriction to the sample span.}
Let
\[
  V=\spn\{y_1,\ldots,y_m\},
\]
and let $P_V$ be the orthogonal projection onto $V$.
For any $y_i\in V$,
\[
  \ip{P_Vz_t}{y_i}=\ip{z_t}{P_Vy_i}=\ip{z_t}{y_i}.
\]
Thus replacing $z_t$ by $P_Vz_t$ leaves every margin and coefficient
unchanged. Projecting \eqref{eq:auxupdate} replaces $e_t$ by $P_Ve_t$,
without increasing its norm or the norm of $z_t$.

We must also put the separator in $V$. The vector $P_Vv^\star$ is nonzero,
since
\[
  \ip{P_Vv^\star}{y_i}=\ip{v^\star}{y_i}\ge\mu>0.
\]
Moreover, orthogonal projection cannot increase a norm, so
$0<\norm{P_Vv^\star}\le1$. Consequently,
\[
  \ip{\frac{P_Vv^\star}{\norm{P_Vv^\star}}}{y_i}
  =\frac{\ip{v^\star}{y_i}}{\norm{P_Vv^\star}}
  \ge\mu.
\]
Replace $v^\star$ by this normalized vector and retain the same notation.
We may now assume
\begin{equation}\label{eq:spanreduction}
  z_t,e_t,v^\star\in V.
\end{equation}
This reduction will be repeated at the beginning of each recursive
application. A separator inherited from a previous projection need not
already belong to the smaller sample span.

\proofstep{Step 2. A preliminary length bound and accumulated error.}
Write
\[
  c_t=\sum_{i=1}^m\lambda_{i,t},
  \qquad a_t=\ip{z_t}{v^\star}.
\]
Taking the inner product of \eqref{eq:auxupdate} with $v^\star$ gives
\[
  a_{t+1}-a_t
  =\sum_i\lambda_{i,t}\ip{y_i}{v^\star}+\ip{e_t}{v^\star}
  \ge\mu c_t-EB^{-6}.
\]
Choose $B$ so that $EB^{-6}\le\mu c/2$.
Since $c_t\ge c$, we obtain
\begin{equation}\label{eq:aincrease}
  a_{t+1}-a_t\ge\frac\mu2c_t\ge\frac{\mu c}{2}
  \qquad(0\le t<T).
\end{equation}
As $a_0=0$, summing \eqref{eq:aincrease} over $0\le t<T-1$ yields
\[
  (T-1)\frac{\mu c}{2}
  \le a_{T-1}\le\norm{z_{T-1}}\le KB.
\]
We have used the norm bound at $T-1$, not at the unbounded endpoint $T$.
It follows that
\begin{equation}\label{eq:rawlength}
  T\le1+\frac{2KB}{\mu c}.
\end{equation}
For $B\ge1$, this implies
\begin{equation}\label{eq:totalerror}
  \sum_{t=0}^{T-1}\norm{e_t}
  \le EB^{-6}\left(1+\frac{2KB}{\mu c}\right)
  \le DB^{-5},
  \qquad
  D=E\left(1+\frac{2K}{\mu c}\right).
\end{equation}
The same bound applies to the sum of error norms on any subinterval.
We will use \eqref{eq:rawlength} only to control these errors, not as our
final estimate for $T$.

\proofstep{Step 3. The case of rank one.}
Suppose $r=1$. By \eqref{eq:spanreduction}, we can write
\[
  y_i=\beta_i v^\star,
  \qquad \mu\le\beta_i\le1,
  \qquad z_t=a_tv^\star.
\]
Using the monotonicity of $h_{i,t}$ and \eqref{eq:aincrease},
\begin{align}
  M_{i,t+1}-M_{i,t}
  &=h_{i,t+1}-h_{i,t}+\beta_i(a_{t+1}-a_t)\notag\\
  &\ge\frac{\mu^2}{2}c_t.
  \label{eq:rankonemargins}
\end{align}
Let $\theta=\mu^2/2$.
For $M\ge\log3$ and $\Delta\ge0$, a direct calculation gives
\begin{align}
  \frac{B/(1+\exp(M+\Delta))}{B/(1+\exp M)}
  &=\exp(-\Delta)
    \frac{1+\exp(-M)}{1+\exp(-(M+\Delta))}\notag\\
  &\le\frac43\exp(-\Delta).
  \label{eq:comparison}
\end{align}
Here $\exp(-M)\le1/3$, and the denominator on the first line is at least one.
For $t<T$, the coefficient condition in \eqref{eq:auxbounds} implies
$M_{i,t}>\log3$. Applying \eqref{eq:comparison} to
\eqref{eq:rankonemargins} and then summing over $i$ yields
\[
  c_{t+1}\le\frac43c_t\exp(-\theta c_t).
\]
The function $s\exp(-\theta s)$ attains its maximum at $s=1/\theta$,
where its value is $1/(\e\theta)$. Thus the first update gives the constant bound
\[
  c_1\le C_0,
  \qquad C_0=\frac4{3\e\theta}.
\]

If $1\le t<T$, summing \eqref{eq:rankonemargins} from time $1$ to time $t-1$
gives
\[
  M_{i,t}-M_{i,1}
  \ge\theta\sum_{j=1}^{t-1}c_j
  \ge\theta c(t-1).
\]
Apply \eqref{eq:comparison} directly between times $1$ and $t$. We obtain
\[
  c\le c_t
  \le\frac43c_1\exp(-\theta c(t-1))
  \le\frac43C_0\exp(-\theta c(t-1)).
\]
The factor $4/3$ appears only once because this is a direct comparison of
two times, rather than a product of successive comparisons.
If $T\ge2$, taking $t=T-1$ shows
\[
  T\le2+
  \frac{\max\{0,\log(4C_0/(3c))\}}{\theta c}.
\]
This also gives a constant upper bound when $T\le1$.
Therefore the auxiliary estimate holds for $r=1$, uniformly in $B$.

\proofstep{Step 4. Growth in the separator direction.}
Assume $m,r\ge2$, and assume the auxiliary estimate has been proved for every
smaller sample count. Set
\[
  A=\frac{16}{\mu},
  \qquad
  z_t=a_tv^\star+\widetilde z_t,
  \qquad \widetilde z_t\perp v^\star.
\]
By \eqref{eq:aincrease}, the number of update indices before $a_t$ first
reaches $A\log B$ is at most
\[
  1+\frac{2A\log B}{\mu c}.
\]
If this level is not reached at any time $t<T$, the same expression bounds
$T$. That is enough, because $r-1\ge1$.
We may therefore restrict attention to times $t<T$ with $a_t\ge A\log B$.
Once this inequality holds, it remains true by \eqref{eq:aincrease}.

At every such time,
\begin{equation}\label{eq:transverse}
  \norm{\widetilde z_t}>\frac\mu4a_t.
\end{equation}
To prove this, suppose instead that $\norm{\widetilde z_t}\le\mu a_t/4$.
Then, for each $i$,
\begin{align*}
  M_{i,t}
  &=b_i+h_{i,t}+a_t\ip{v^\star}{y_i}
      +\ip{\widetilde z_t}{y_i}\\
  &\ge\log3+\mu a_t-\norm{\widetilde z_t}\\
  &\ge\log3+\frac{3\mu}{4}a_t.
\end{align*}
We used $a_t\ge0$, $h_{i,t}\ge0$, and $\norm{y_i}\le1$.
Since $1+\exp M\ge\exp M$, this implies
\[
  \sum_i\lambda_{i,t}
  \le\frac{mB}{3}\exp\left(-\frac{3\mu}{4}a_t\right)
  \le\frac m3 B^{1-3\mu A/4}
  =\frac m3B^{-11}.
\]
For sufficiently large $B$, the final expression is smaller than $c$,
contradicting \eqref{eq:auxbounds}. This proves \eqref{eq:transverse}.
Geometrically, if the coefficient mass remains at least $c$, the trajectory
cannot point too closely in the separator direction, since that would make
every margin large.

\proofstep{Step 5. Choosing a subset during one phase.}
Fix a time $s<T$ with $a_s\ge A\log B$. Define
\[
  R=\norm{\widetilde z_s},
  \qquad u=\frac{\widetilde z_s}{R},
  \qquad \kappa=\frac{\mu a_s}{4R}.
\]
The vector $u$ and the numbers $R,\kappa$ remain fixed during this phase.
By \eqref{eq:spanreduction} and \eqref{eq:transverse},
\begin{equation}\label{eq:udirection}
  u\in V,
  \qquad \norm{u}=1,
  \qquad u\perp v^\star,
  \qquad 0<\kappa<1.
\end{equation}
In particular, $u\in V$ uses the fact that both $z_s$ and $v^\star$ belong
to $V$.
For each sample, set
\[
  \alpha_i=\ip{u}{y_i},
  \qquad S=\{i:\alpha_i\le-\kappa\}.
\]
The inequality $\norm{y_i}\le1$ gives $|\alpha_i|\le1$.
The set $S$ contains the samples that have a sufficiently negative component
along $u$.

For $t\ge s$, decompose
\begin{equation}\label{eq:decomposition}
  z_t=a_tv^\star+\rho_tu+\zeta_t,
  \qquad \rho_t=\ip{z_t}{u},
  \qquad \zeta_t\perp v^\star,
  \qquad \zeta_t\perp u.
\end{equation}
At the start, $\rho_s=R$ and $\zeta_s=0$.
Let $\sigma$ be the first time in $\{s+1,\ldots,T-1\}$ at which
\begin{equation}\label{eq:phasestop}
  \rho_\sigma\le0
  \quad\text{or}\quad
  \norm{\zeta_\sigma}\ge\frac\mu8a_\sigma.
\end{equation}
If no such time exists, set $\sigma=T$.
The phase contains the update indices $s\le t<\sigma$.
Thus it always contains at least one update, and before its endpoint we have
\[
  \rho_t>0,
  \qquad \norm{\zeta_t}<\frac\mu8a_t.
\]

We next show that $S$ is nonempty and proper.
If $S$ were empty, then $\alpha_i>-\kappa$ for every $i$, and at time $s$,
\[
  M_{i,s}
  =b_i+h_{i,s}+a_s\ip{v^\star}{y_i}+R\alpha_i
  >\log3+\mu a_s-\kappa R
  =\log3+\frac{3\mu}{4}a_s.
\]
As in Step 4, this would give total coefficient mass smaller than $c$.
Therefore $S$ is nonempty.

To show that $S$ is proper, sum the exact recursion from the original
initial time $0$ to time $s$. Since $z_0=0$,
\[
  z_s=\sum_{j=0}^{s-1}\sum_i\lambda_{i,j}y_i+\mathcal E_s,
  \qquad
  \mathcal E_s=\sum_{j=0}^{s-1}e_j,
  \qquad \norm{\mathcal E_s}\le DB^{-5}.
\]
If every sample belonged to $S$, then every $\alpha_i\le-\kappa<0$.
Taking the inner product with $u$ would yield
\[
  R=\sum_{j=0}^{s-1}\sum_i\lambda_{i,j}\alpha_i
     +\ip{\mathcal E_s}{u}
  \le DB^{-5},
\]
because all coefficients are positive. But \eqref{eq:transverse} gives
\[
  R>\frac\mu4a_s\ge\frac{\mu A}{4}\log B=4\log B.
\]
These two bounds are incompatible for sufficiently large $B$.
We have proved
\begin{equation}\label{eq:properS}
  1\le|S|\le m-1.
\end{equation}

\proofstep{Step 6. Small coefficients outside the subset.}
We prove, by induction over the update indices in the phase, that
\begin{equation}\label{eq:rhomonotone}
  0<\rho_t\le R\quad(s\le t<\sigma),
  \qquad
  \rho_{t+1}<\rho_t\quad(s\le t<\sigma).
\end{equation}
The first upper bound holds at $t=s$ because $\rho_s=R$.
Suppose it has been proved at the current time $t<\sigma$.
The stopping rule gives $\rho_t>0$.
For $i\notin S$, we have $\alpha_i>-\kappa$, and hence
\[
  \alpha_i\rho_t\ge-\kappa\rho_t\ge-\kappa R.
\]
Using \eqref{eq:decomposition} and the stopping rule,
\begin{align*}
  M_{i,t}
  &\ge\log3+\mu a_t+\alpha_i\rho_t-\norm{\zeta_t}\\
  &\ge\log3+\frac{7\mu}{8}a_t-\kappa R\\
  &=\log3+\frac{7\mu}{8}a_t-\frac\mu4a_s\\
  &\ge\log3+\frac{5\mu}{8}a_t.
\end{align*}
The last inequality follows from $a_t\ge a_s$, which is a consequence of
\eqref{eq:aincrease}. Therefore
\begin{equation}\label{eq:outsideS}
  \sum_{i\notin S}\lambda_{i,t}
  \le\frac{mB}{3}\exp\left(-\frac{5\mu}{8}a_t\right)
  \le\frac m3B^{1-5\mu A/8}
  =\frac m3B^{-9}
  \le B^{-6}
\end{equation}
for sufficiently large $B$.
Subtracting this from the total mass bound gives
\begin{equation}\label{eq:insideS}
  \sum_{i\in S}\lambda_{i,t}\ge c-B^{-6}\ge\frac c2.
\end{equation}
Taking the inner product of the update with $u$, we now obtain
\begin{align*}
  \rho_{t+1}-\rho_t
  &=\sum_{i\in S}\alpha_i\lambda_{i,t}
    +\sum_{i\notin S}\alpha_i\lambda_{i,t}+\ip{e_t}{u}\\
  &\le-\kappa\sum_{i\in S}\lambda_{i,t}
       +\sum_{i\notin S}\lambda_{i,t}+\norm{e_t}\\
  &\le-\frac{\kappa c}{2}+(E+1)B^{-6}.
\end{align*}
We must compare the error with $\kappa$, since $\kappa$ itself may depend on
$B$. Orthogonality in $z_s=a_sv^\star+\widetilde z_s$ gives
$R\le\norm{z_s}\le KB$. Thus
\begin{equation}\label{eq:kappalower}
  \kappa=\frac{\mu a_s}{4R}
  \ge\frac{\mu A\log B}{4KB}.
\end{equation}
Choose $B$ so that
\[
  (E+1)B^{-6}\le\frac{\mu A c\log B}{16KB}.
\]
Such a threshold depends only on the fixed scalar parameters, because
$B^5\log B$ tends to infinity. The right side is at most $\kappa c/4$ by
\eqref{eq:kappalower}. Consequently,
\[
  \rho_{t+1}-\rho_t\le-\frac{\kappa c}{4}<0.
\]
If $t+1<\sigma$, positivity of $\rho_{t+1}$ again follows from the stopping
rule, and the induction continues. If $t+1=\sigma$, the strict decrease
still holds, although $\rho_\sigma$ may be nonpositive.
This proves \eqref{eq:rhomonotone}, \eqref{eq:outsideS}, and
\eqref{eq:insideS} for every update in the phase, including its last update.

\proofstep{Step 7. The exact projected process.}
Let
\[
  P=I-uu^\top.
\]
This is the orthogonal projection onto $u^\perp$.
For $i\in S$ and $s\le t\le\sigma$, define
\begin{align*}
  y_i'&=Py_i,
  & q_t&=P(z_t-z_s),\\
  b_i'&=M_{i,s},
  & h_{i,t}'&=h_{i,t}-h_{i,s}+\alpha_i(\rho_t-R).
\end{align*}
We verify each hypothesis of the auxiliary estimate for these new quantities.

First, $\norm{y_i'}\le\norm{y_i}\le1$.
Since $u\perp v^\star$, we have $Pv^\star=v^\star$, so
\[
  \ip{v^\star}{y_i'}
  =\ip{Pv^\star}{y_i}
  =\ip{v^\star}{y_i}\ge\mu.
\]
Thus the same unit separator is still valid.
It may not belong to the span of the retained samples, but Step 1 will
address this at the start of the recursive application.

Second, the rank decreases. On the space $V$, the kernel of $P$ is exactly
$\spn\{u\}$. Indeed, $Px=0$ is equivalent to $x=\ip{u}{x}u$, and
$u\in V$ by \eqref{eq:udirection}.
The dimension formula for a linear map therefore gives
$\dim P(V)=r-1$. Since every $y_i'$ belongs to $P(V)$,
\begin{equation}\label{eq:rankdrop}
  1\le r':=\dim\spn\{y_i':i\in S\}\le r-1.
\end{equation}
The lower bound follows from $S\ne\varnothing$ and
$\ip{v^\star}{y_i'}\ge\mu>0$.

Third, the new process starts at zero, since $q_s=0$.
The original coefficient bound at time $s$ gives
\[
  \frac{B}{1+\exp(M_{i,s})}<\frac B4
  \quad\Longrightarrow\quad M_{i,s}>\log3.
\]
Hence $b_i'>\log3$.
We also have $h_{i,s}'=0$, and, for $s\le t<\sigma$,
\[
  h_{i,t+1}'-h_{i,t}'
  =h_{i,t+1}-h_{i,t}
   +\alpha_i(\rho_{t+1}-\rho_t)\ge0.
\]
The first term is nonnegative by assumption. The product is nonnegative
because both $\alpha_i$ and $\rho_{t+1}-\rho_t$ are negative.
This remains true for the update ending at $\sigma$.
Thus the new offsets are nonnegative and nondecreasing through the endpoint.

Fourth, the margins retain their exact logistic representation.
By the definition of $q_t$,
\[
  z_t-z_s=q_t+(\rho_t-R)u.
\]
Since $q_t\perp u$, we have $\ip{q_t}{y_i}=\ip{q_t}{Py_i}$.
It follows that
\begin{align}
  M_{i,t}
  &=M_{i,s}+h_{i,t}-h_{i,s}+\ip{z_t-z_s}{y_i}\notag\\
  &=M_{i,s}+h_{i,t}-h_{i,s}
       +\alpha_i(\rho_t-R)+\ip{q_t}{Py_i}\notag\\
  &=b_i'+h_{i,t}'+\ip{q_t}{y_i'}.
  \label{eq:projectedmargin}
\end{align}
In particular, the coefficient of $y_i'$ is the original $\lambda_{i,t}$,
with the same $B$.

Finally, projecting the update gives
\begin{align}
  q_{t+1}-q_t
  &=P(z_{t+1}-z_t)\notag\\
  &=\sum_{i\in S}\lambda_{i,t}y_i'+e_t',
  \qquad
  e_t'=Pe_t+\sum_{i\notin S}\lambda_{i,t}Py_i.
  \label{eq:projectedupdate}
\end{align}
By the triangle inequality and \eqref{eq:outsideS},
\[
  \norm{e_t'}
  \le\norm{e_t}+\sum_{i\notin S}\lambda_{i,t}\norm{Py_i}
  \le(E+1)B^{-6}.
\]
For $s\le t<\sigma$, we also have
\[
  \sum_{i\in S}\lambda_{i,t}\ge\frac c2,
  \qquad
  \norm{q_t}\le\norm{z_t-z_s}
  \le\norm{z_t}+\norm{z_s}\le2KB.
\]
Every retained coefficient is still less than $B/4$.
After shifting the time origin from $0$ to $s$ and relabeling the retained
samples, this process satisfies all assumptions of the auxiliary estimate with
\begin{equation}\label{eq:recursiveparameters}
  (m,\mu,c,K,E)
  \quad\text{replaced by}\quad
  (|S|,\mu,c/2,2K,E+1).
\end{equation}
The number of samples is smaller by \eqref{eq:properS}.
By the induction hypothesis and \eqref{eq:rankdrop},
\begin{equation}\label{eq:phaselength}
  \sigma-s
  \le C(1+\log B)^{r'-1}
  \le C(1+\log B)^{r-2}.
\end{equation}
To make the constant and the threshold uniform over all possible $S$, take
the maximum of the inductive constants for sample counts $1,\ldots,m-1$
with the scalar parameters in \eqref{eq:recursiveparameters}.
There are only finitely many such counts.
The estimate does not require $q_\sigma$ or the coefficients at time
$\sigma$ to satisfy the interior bounds.

\proofstep{Step 8. Counting the phases.}
The preceding estimate controls one phase. We now show that there can be
only a logarithmic number of phases.
Suppose $\sigma<T$, so that at least one event in \eqref{eq:phasestop}
has occurred. Set $\Delta z=z_\sigma-z_s$.
Summing the update over the phase gives
\[
  \Delta z=g+\mathcal E,
  \qquad g=\sum_i A_i y_i,
  \qquad A_i=\sum_{t=s}^{\sigma-1}\lambda_{i,t}\ge0,
  \qquad \norm{\mathcal E}\le DB^{-5}.
\]
The positive combination $g$ satisfies
\[
  \norm{g}\le\sum_i A_i\norm{y_i}\le\sum_i A_i,
  \qquad
  \ip{v^\star}{g}\ge\mu\sum_i A_i\ge\mu\norm{g}.
\]
Using $\norm{g}\ge\norm{\Delta z}-\norm{\mathcal E}$, we obtain
\begin{align}
  a_\sigma-a_s
  &=\ip{v^\star}{\Delta z}\notag\\
  &\ge\mu\norm{g}-\norm{\mathcal E}\notag\\
  &\ge\mu\norm{\Delta z}-(1+\mu)\norm{\mathcal E}\notag\\
  &\ge\mu\norm{\Delta z}-2DB^{-5}.
  \label{eq:coneincrement}
\end{align}
Here $\mu\le1$ was used in the last line.
Thus a large displacement necessarily produces a substantial increase in
the separator component.

Choose $B$ large enough that
\begin{equation}\label{eq:absorption}
  2DB^{-5}\le\frac{\mu^2 A\log B}{16}.
\end{equation}
This threshold depends only on the fixed parameters.
If $\norm{\zeta_\sigma}\ge\mu a_\sigma/8$, then the orthogonal decomposition
\[
  \Delta z=(a_\sigma-a_s)v^\star+(\rho_\sigma-R)u+\zeta_\sigma
\]
implies $\norm{\Delta z}\ge\norm{\zeta_\sigma}$.
Since $a_\sigma\ge a_s\ge A\log B$, equations
\eqref{eq:coneincrement} and \eqref{eq:absorption} yield
\[
  a_\sigma-a_s
  \ge\frac{\mu^2}{8}a_\sigma-2DB^{-5}
  \ge\frac{\mu^2}{16}a_\sigma.
\]
Rearranging gives
\[
  a_\sigma\ge\frac{a_s}{1-\mu^2/16}.
\]
If instead $\rho_\sigma\le0$, then
\[
  \norm{\Delta z}\ge|\rho_\sigma-R|\ge R>\frac\mu4a_s.
\]
The same two estimates imply
\[
  a_\sigma-a_s
  \ge\frac{\mu^2}{4}a_s-2DB^{-5}
  \ge\frac{\mu^2}{8}a_s.
\]
Since $0<\mu\le1$ and $(1-x)^{-1}\ge1+x$ for $0\le x<1$, both cases give
\begin{equation}\label{eq:multiplicative}
  a_\sigma\ge\left(1+\frac{\mu^2}{16}\right)a_s.
\end{equation}

At time $\sigma<T$, start a new phase for the same original process,
choosing a new $u$ and a new $S$ from its current state.
The initial condition used in the positive combination argument remains
$z_0=0$. The time shift is used only for the recursive process of Step 7.
Since $a_t$ is increasing, every new phase begins with $a_t\ge A\log B$.

Let $N$ be the number of phases whose endpoints are strictly smaller than
$T$, and let $s_0$ be the start of the first phase.
After these $N$ phases, repeated use of \eqref{eq:multiplicative} gives
\[
  \left(1+\frac{\mu^2}{16}\right)^N A\log B
  \le a_{s_N}\le\norm{z_{s_N}}\le KB,
\]
where $s_N<T$ is the endpoint of the last such phase.
For $N=0$, set $s_N=s_0$.
Taking logarithms yields
\[
  N\le
  \frac{\log\bigl(KB/(A\log B)\bigr)}
       {\log(1+\mu^2/16)}
  \le C(1+\log B).
\]
The argument $KB/(A\log B)$ of the logarithm is at least one whenever
a phase exists, because then $a_{s_0}\ge A\log B$ and $a_{s_0}\le KB$.
There is at most one additional phase ending at $T$.

\proofstep{Step 9. Completion of the induction.}
The initial part before $a_t\ge A\log B$ has length at most
$C(1+\log B)$. There are at most $C(1+\log B)$ subsequent phases, and
\eqref{eq:phaselength} bounds each one. Therefore
\begin{align*}
  T
  &\le C(1+\log B)
       +C(1+\log B)(1+\log B)^{r-2}\\
  &\le C(1+\log B)^{r-1},
\end{align*}
since $r\ge2$. This proves the inductive step.

For completeness, the dependence of the constants can be followed through
the recursion. Every recursive application removes at least one sample,
so there are at most $m-1$ reductions along any sequence of applications.
The margin lower bound stays at least $\mu$, including after the restriction
to the new sample span in Step 1.
After $j$ reductions, the remaining scalar parameters have the form
\[
  \frac{c}{2^j},
  \qquad 2^jK,
  \qquad E+j,
  \qquad 0\le j\le m-1.
\]
Only finitely many sample counts and these finitely many parameter choices
can occur. Taking maxima over them gives constants depending only on
$m,\mu,c,K,E$. No geometric quantity other than the norm and margin bounds
enters the estimates. In particular, the constants do not depend on rank
or ambient dimension.

This proves the precise form of Lemma~\ref{lem:sketch-recursion} used
below. The fixed offsets, initialization, coefficient threshold, and
error bound in that statement are specified in
\eqref{eq:auxmargin} through \eqref{eq:auxbounds}.

\subsection{The original gradient trajectory}\label{sub:upperoriginal}
We return to the original data $x_1,\ldots,x_n$ and set
\[
 r=\dim\spn\{x_1,\ldots,x_n\}.
\]
Here $r$ again denotes the rank of the original data, rather than the rank
of the preceding auxiliary process. We will obtain the stronger estimate
\[
 \tau_\eta\le C_{n,\gamma}(1+\log\eta)^{\max\{r-2,0\}}.
\]
Since $1\le r\le\min\{n,d\}$, this implies \eqref{eq:appendixupper}.

Throughout this proof, the constants and all lower thresholds on $\eta$
depend only on $n$ and $\gamma$.
We take $\eta$ large enough that $B=\eta/n\ge\e$.
Every additional threshold imposed below will have the same permitted
dependence.

\proofstep{Step 1. The exact coefficients and an entropy inequality.}
The derivative of the scalar logistic loss is
\[
  \frac{\mathrm d}{\mathrm d s}\log(1+\exp(-s))
  =-\frac{\exp(-s)}{1+\exp(-s)}
  =-\frac1{1+\exp s}.
\]
Thus, with
\[
  B=\frac\eta n,
  \qquad m_{i,t}=\ip{w_t}{x_i},
  \qquad \lambda_{i,t}=\frac{B}{1+\exp(m_{i,t})},
  \qquad c_t=\sum_i\lambda_{i,t},
\]
the gradient descent update becomes
\begin{equation}\label{eq:originalupdate}
  w_{t+1}=w_t+\sum_i\lambda_{i,t}x_i=w_t+c_tu_t,
  \qquad
  u_t=\frac1{c_t}\sum_i\lambda_{i,t}x_i.
\end{equation}
Every coefficient is strictly positive, so $c_t>0$ and $u_t$ is well defined.
As a convex combination of the samples, $u_t$ satisfies
\begin{equation}\label{eq:convexbounds}
  \norm{u_t}\le1,
  \qquad \ip{w^\star}{u_t}\ge\gamma,
  \qquad 0<c_t\le nB=\eta.
\end{equation}

Set $\pi_{i,t}=\lambda_{i,t}/c_t$, so that $\pi_{i,t}>0$ and
$\sum_i \pi_{i,t}=1$.
The coefficient formula gives
\[
  m_{i,t}=\log\frac{B-\lambda_{i,t}}{\lambda_{i,t}}
  \le\log\frac{B}{\lambda_{i,t}}.
\]
Consequently,
\begin{align*}
  \ip{w_t}{u_t}
  &=\sum_i \pi_{i,t}m_{i,t}\\
  &\le\sum_i \pi_{i,t}\log\frac{B}{c_t \pi_{i,t}}\\
  &=\log\frac{B}{c_t}+\sum_i \pi_{i,t}\log\frac1{\pi_{i,t}}.
\end{align*}
The last sum is bounded by $\log n$. Indeed, concavity of the logarithm
and the fact that the $\pi_{i,t}$ sum to one give
\[
  \sum_i \pi_{i,t}\log\frac1{\pi_{i,t}}
  \le\log\left(\sum_i \pi_{i,t}\frac1{\pi_{i,t}}\right)
  =\log n.
\]
We have proved
\begin{equation}\label{eq:entropy}
  \ip{w_t}{u_t}\le\log\frac{nB}{c_t}
  =\log\frac\eta{c_t}.
\end{equation}
This inequality relates the current parameter to the direction of its
next update without requiring any bound on the individual margins.

\proofstep{Step 2. Finite transition time and total coefficient mass.}
We first obtain a lower bound on $c_t$ before the transition.
The identity
\[
  1+\exp(-m_{i,t})=\frac1{1-\lambda_{i,t}/B}
\]
implies
\begin{equation}\label{eq:losscorrelation}
  \eta F(w_t)=B\sum_i-\log(1-\lambda_{i,t}/B).
\end{equation}
For $0\le q\le1/4$, integration gives
\[
  -\log(1-q)=\int_0^q\frac{\mathrm ds}{1-s}
  \le\frac{q}{1-q}\le\frac43q.
\]
If every $\lambda_{i,t}<B/4$, applying this inequality in
\eqref{eq:losscorrelation} yields
\[
  \eta F(w_t)\le\frac43c_t.
\]
When $t<\tau_\eta$, the definition of $\tau_\eta$ gives
$\eta F(w_t)>1/8$, so $c_t>3/32$.
If some coefficient is at least $B/4$, then instead
$c_t\ge B/4\ge3/32$, where the last inequality holds because $B\ge\e$.
Thus, in both cases,
\begin{equation}\label{eq:masslower}
  c_t\ge\frac3{32}\qquad(t<\tau_\eta).
\end{equation}

Consider the squared distance
\[
  Z_t=\left\|w_t-\frac{2\eta}{\gamma}w^\star\right\|^2.
\]
Using \eqref{eq:originalupdate}, expanding the square, and then applying
\eqref{eq:convexbounds} and \eqref{eq:entropy}, we get
\begin{align*}
  Z_{t+1}-Z_t
  &=2c_t\left\langle w_t-\frac{2\eta}{\gamma}w^\star,u_t\right\rangle
      +c_t^2\norm{u_t}^2\\
  &\le2c_t\left(\log\frac\eta{c_t}-2\eta\right)+c_t^2\\
  &\le2c_t\log\frac\eta{c_t}-3\eta c_t.
\end{align*}
The last step uses $c_t^2\le\eta c_t$.
Choose $\eta$ sufficiently large that
\[
  2\log\frac{32\eta}{3}\le\eta.
\]
Such a threshold exists since $\log\eta/\eta$ tends to zero.
For $t<\tau_\eta$, \eqref{eq:masslower} then gives
\begin{equation}\label{eq:potentialdecay}
  Z_{t+1}-Z_t\le-2\eta c_t.
\end{equation}

We have not yet proved that $\tau_\eta$ is finite, so we sum only over a
finite prefix. For any finite integer $N\le\tau_\eta$,
\[
  Z_N-Z_0
  =\sum_{t=0}^{N-1}(Z_{t+1}-Z_t)
  \le-2\eta\sum_{t=0}^{N-1}c_t.
\]
Since $Z_N\ge0$ and $Z_0=4\eta^2/\gamma^2$, this gives
\begin{equation}\label{eq:finiteprefix}
  \sum_{t=0}^{N-1}c_t\le\frac{2\eta}{\gamma^2},
  \qquad Z_N\le Z_0.
\end{equation}
If $\tau_\eta=\infty$, then \eqref{eq:masslower} and
\eqref{eq:finiteprefix} would imply
\[
  \frac{3N}{32}\le\frac{2\eta}{\gamma^2}
\]
for every positive integer $N$. This is impossible.
Thus $\tau_\eta<\infty$, and we may now set $N=\tau_\eta$ in
\eqref{eq:finiteprefix} to obtain
\begin{equation}\label{eq:globalmass}
  \sum_{t<\tau_\eta}c_t\le\frac{2\eta}{\gamma^2}.
\end{equation}
For every $0\le t\le\tau_\eta$, the second part of
\eqref{eq:finiteprefix} and the triangle inequality also give
\begin{equation}\label{eq:globalnorm}
  \norm{w_t}
  \le\left\|w_t-\frac{2\eta}{\gamma}w^\star\right\|
       +\frac{2\eta}{\gamma}
  \le\frac{4\eta}{\gamma}
  =\frac{4nB}{\gamma}.
\end{equation}

Equations~\eqref{eq:masslower}, \eqref{eq:globalmass}, and
\eqref{eq:globalnorm} prove Lemma~\ref{lem:sketch-mass}, with the
universal choice $c_0=3/32$.

Let $N_{\mathrm{large}}$ be the number of update indices $t<\tau_\eta$
for which some $\lambda_{i,t}\ge B/4$.
At each such index $c_t\ge B/4$, so \eqref{eq:globalmass} yields
\begin{equation}\label{eq:largecount}
  N_{\mathrm{large}}\frac B4
  \le\sum_{t<\tau_\eta}c_t
  \le\frac{2\eta}{\gamma^2},
  \qquad
  N_{\mathrm{large}}\le\frac{8n}{\gamma^2}.
\end{equation}
Thus only a bounded number of updates have a coefficient above the threshold.

\proofstep{Step 3. The case of rank one.}
Suppose the original data have rank $r=1$.
Choose a unit vector $v$ along their common line and orient it so that
$\ip{w^\star}{v}>0$.
Write $x_i=a_i v$. The separator condition gives
\[
  a_i\ip{w^\star}{v}\ge\gamma.
\]
Because $0<\ip{w^\star}{v}\le1$, we have $a_i\ge\gamma$ for every $i$.
At the origin, every logistic coefficient equals $B/2$. Therefore
\[
  w_1=\frac B2\sum_jx_j,
  \qquad
  \ip{w_1}{x_i}
  =\frac B2a_i\sum_ja_j
  \ge\frac{nB\gamma^2}{2}
  =\frac{\eta\gamma^2}{2}.
\]
Using $\log(1+s)\le s$ for $s\ge0$,
\[
  F(w_1)
  \le\frac1n\sum_i\exp(-\ip{w_1}{x_i})
  \le\exp\left(-\frac{\eta\gamma^2}{2}\right).
\]
For sufficiently large $\eta$, the last expression is at most $1/(8\eta)$.
Hence $\tau_\eta\le1$, which proves \eqref{eq:appendixupper} when $r=1$.
For the rest of the proof, assume $r\ge2$.

\proofstep{Step 4. A projection at the start of an interval.}
The remaining update indices are those satisfying
\begin{equation}\label{eq:smallcondition}
  \lambda_{i,t}<\frac B4\qquad\text{for every }i.
\end{equation}
Partition them into maximal consecutive integer intervals $[a,b)$ within
$\{0,\ldots,\tau_\eta-1\}$.
Here $[a,b)$ means the indices $a,a+1,\ldots,b-1$, so that it contains
$b-a$ updates and ends at the state $w_b$.
There are at most $N_{\mathrm{large}}+1$ such intervals.
Their endpoints may be a time with a large coefficient or the transition
time itself.

Fix one of these intervals. We will partition it further into blocks.
Let $s\in[a,b)$ be the start of the current block, initially $s=a$.
Every new block will use a newly defined projection.
Since $\lambda_{i,0}=B/2$, time $0$ does not satisfy
\eqref{eq:smallcondition}. Thus $s\ge1$.
Summing the updates from time $0$ gives
\begin{equation}\label{eq:positivecombination}
  w_s=\sum_i A_i x_i,
  \qquad A_i=\sum_{t=0}^{s-1}\lambda_{i,t}>0.
\end{equation}
The separator component is increasing, because
\[
  \ip{w_{t+1}-w_t}{w^\star}
  =\sum_i\lambda_{i,t}\ip{x_i}{w^\star}
  \ge\gamma c_t>0.
\]
The first update consequently gives
\[
  \ip{w_s}{w^\star}
  \ge\ip{w_1}{w^\star}
  =\frac B2\sum_i\ip{x_i}{w^\star}
  \ge\frac{\eta\gamma}{2}.
\]
Set
\[
  R=\norm{w_s},
  \qquad u=\frac{w_s}{R},
  \qquad P=I-uu^\top.
\]
These quantities are well defined and satisfy
\begin{equation}\label{eq:radiallower}
  R\ge\frac{\eta\gamma}{2},
  \qquad
  0<\ip{w^\star}{u}\le1.
\end{equation}

Define the subset
\begin{equation}\label{eq:G}
  G=\left\{i:\ip{Pw^\star}{Px_i}\ge\frac\gamma2\right\}.
\end{equation}
It is proper. Indeed, $Pw_s=0$, so \eqref{eq:positivecombination} gives
\[
  0=\ip{Pw^\star}{Pw_s}
   =\sum_i A_i\ip{Pw^\star}{Px_i}.
\]
If every sample belonged to $G$, the last sum would be at least
$(\gamma/2)\sum_i A_i>0$, a contradiction.
Therefore
\begin{equation}\label{eq:Gproper}
  |G|\le n-1.
\end{equation}

Put $\alpha_i=\ip{u}{x_i}$. The small coefficient condition at time $s$ is
equivalent to $m_{i,s}>\log3$, and hence
\begin{equation}\label{eq:alphapositive}
  \alpha_i=\frac{m_{i,s}}R>0
  \qquad\text{for every }i.
\end{equation}
For $i\notin G$, the projection identity gives
\[
  \ip{w^\star}{u}\alpha_i
  =\ip{w^\star}{x_i}-\ip{Pw^\star}{Px_i}
  >\gamma-\frac\gamma2=\frac\gamma2.
\]
By \eqref{eq:radiallower}, $0<\ip{w^\star}{u}\le1$.
Dividing by this positive number shows $\alpha_i>\gamma/2$.
Thus every sample outside $G$ starts the block with margin
\begin{equation}\label{eq:outsidemargin}
  m_{i,s}=R\alpha_i\ge\frac{\eta\gamma^2}{4}
  \qquad(i\notin G).
\end{equation}
This bound is proportional to $\eta$, rather than to $\log\eta$.
It will limit how many times an outside sample can become relevant again.

\proofstep{Step 5. Defining each block and bounding the number of blocks.}
Increase the threshold on $\eta$ so that
\begin{equation}\label{eq:returnthreshold}
  8\log B\le\frac{\eta\gamma^2}{8}.
\end{equation}
For the fixed $u,P,G$ chosen at time $s$, let $\sigma$ be the first time
in $\{s+1,\ldots,b-1\}$ at which some $i\notin G$ satisfies
$m_{i,\sigma}\le8\log B$. If no such time exists, set $\sigma=b$.
The block is the update interval $[s,\sigma)$.
Its initial margins satisfy $m_{i,s}>8\log B$ for $i\notin G$ by
\eqref{eq:outsidemargin} and \eqref{eq:returnthreshold}.
Consequently, at every update index in the block,
\begin{equation}\label{eq:outeroutside}
  \lambda_{i,t}
  \le B\exp(-m_{i,t})
  \le B\exp(-8\log B)
  =B^{-7}
  \qquad(i\notin G).
\end{equation}

If $\sigma<b$, a new block begins at time $\sigma$.
We then repeat Step 4 using this new start, recomputing $R,u,P$, and $G$.
In particular, \eqref{eq:outsidemargin} holds for the outside samples of
this new subset. We do not reuse the preceding projection or subset.
Because each block has at least one update and $b$ is finite, this procedure
partitions the entire interval.

We now count all blocks, across all maximal intervals satisfying
\eqref{eq:smallcondition}.
For every sample and every time,
\begin{align*}
  |m_{i,t+1}-m_{i,t}|
  &=\left|\sum_j\lambda_{j,t}\ip{x_i}{x_j}\right|\\
  &\le\sum_j\lambda_{j,t}|\ip{x_i}{x_j}|\\
  &\le\sum_j\lambda_{j,t}=c_t.
\end{align*}
The last step uses $|\ip{x_i}{x_j}|\le\norm{x_i}\norm{x_j}\le1$.
If a block ends at $\sigma<b$, choose an outside sample with
$m_{i,\sigma}\le8\log B$.
By telescoping its margin changes and using
\eqref{eq:outsidemargin} and \eqref{eq:returnthreshold},
\begin{align*}
  \sum_{t=s}^{\sigma-1}c_t
  &\ge\sum_{t=s}^{\sigma-1}|m_{i,t+1}-m_{i,t}|\\
  &\ge|m_{i,\sigma}-m_{i,s}|\\
  &\ge m_{i,s}-m_{i,\sigma}\\
  &\ge\frac{\eta\gamma^2}{4}-8\log B
   \ge\frac{\eta\gamma^2}{8}.
\end{align*}
Thus every such endpoint consumes at least $\eta\gamma^2/8$ of the global
coefficient mass.

Let $N_{\mathrm{return}}$ be the number of block endpoints lying strictly
inside their maximal intervals. All corresponding update intervals are
disjoint, including those in different maximal intervals.
The mass bound \eqref{eq:globalmass} therefore gives
\begin{equation}\label{eq:returncount}
  N_{\mathrm{return}}\frac{\eta\gamma^2}{8}
  \le\frac{2\eta}{\gamma^2},
  \qquad
  N_{\mathrm{return}}\le\frac{16}{\gamma^4}.
\end{equation}
Each maximal interval contributes one initial block, and each interior
endpoint contributes one additional block. Hence the total number satisfies
\begin{equation}\label{eq:blockcount}
  N_{\mathrm{blocks}}
  \le N_{\mathrm{large}}+1+N_{\mathrm{return}}
  \le\frac{8n}{\gamma^2}+1+\frac{16}{\gamma^4}.
\end{equation}
This uses one mass budget for the whole trajectory, not a separate budget
for each interval.

\proofstep{Step 6. Applying the auxiliary estimate to one block.}
Fix a block $[s,\sigma)$, with the quantities $R,u,P,G$ recomputed at its
own initial time $s$.
For $s\le t\le\sigma$, write
\[
  w_t=\rho_tu+z_t,
  \qquad \rho_t=\ip{w_t}{u},
  \qquad z_t=Pw_t.
\]
Since $w_s=Ru$, we have $\rho_s=R$ and $z_s=0$.
Taking the inner product of \eqref{eq:originalupdate} with $u$, and using
\eqref{eq:alphapositive}, gives
\begin{equation}\label{eq:outermonotonicity}
  \rho_{t+1}-\rho_t=\sum_i\lambda_{i,t}\alpha_i\ge0.
\end{equation}
The $\alpha_i$ are fixed by the block's initial direction, so their
positivity remains valid throughout the block.

For $i\in G$, define
\[
  y_i=Px_i,
  \qquad b_i=m_{i,s}=R\alpha_i,
  \qquad h_{i,t}=\alpha_i(\rho_t-R).
\]
Then $\norm{y_i}\le1$, $b_i>\log3$, and $h_{i,s}=0$.
By \eqref{eq:outermonotonicity} and $\alpha_i>0$, the offsets are nonnegative
and nondecreasing, including through time $\sigma$.
Since $z_t\perp u$ and $x_i=\alpha_i u+y_i$,
\begin{align}
  m_{i,t}
  &=\ip{\rho_tu+z_t}{\alpha_i u+y_i}\notag\\
  &=\rho_t\alpha_i+\ip{z_t}{y_i}\notag\\
  &=b_i+h_{i,t}+\ip{z_t}{y_i}.
  \label{eq:outermarginidentity}
\end{align}
This is exactly the auxiliary margin formula.
Projecting \eqref{eq:originalupdate} also gives
\begin{equation}\label{eq:outerupdate}
  z_{t+1}=z_t+\sum_{i\in G}\lambda_{i,t}y_i+e_t,
  \qquad e_t=\sum_{i\notin G}\lambda_{i,t}Px_i.
\end{equation}
Using \eqref{eq:outeroutside},
\[
  \norm{e_t}
  \le\sum_{i\notin G}\lambda_{i,t}\norm{Px_i}
  \le nB^{-7}\le B^{-6}
\]
for $B\ge n$.
The global lower bound \eqref{eq:masslower} gives
\begin{equation}\label{eq:outermass}
  \sum_{i\in G}\lambda_{i,t}
  =c_t-\sum_{i\notin G}\lambda_{i,t}
  \ge\frac3{32}-nB^{-7}
  \ge\frac3{64}
  \qquad(s\le t<\sigma)
\end{equation}
for sufficiently large $B$.
In particular, $G$ is nonempty.
Together with \eqref{eq:Gproper}, this proves $1\le|G|\le n-1$.

Nonemptiness of $G$ also implies $Pw^\star\ne0$, by its definition in
\eqref{eq:G}. Thus the unit vector
\[
  v^\star=\frac{Pw^\star}{\norm{Pw^\star}}
\]
satisfies
\[
  \ip{v^\star}{y_i}
  =\frac{\ip{Pw^\star}{Px_i}}{\norm{Pw^\star}}
  \ge\frac\gamma2
  \qquad(i\in G),
\]
where we used $\norm{Pw^\star}\le1$.
The norm bound \eqref{eq:globalnorm} yields
\[
  \norm{z_t}\le\norm{w_t}\le\frac{4nB}{\gamma}
  \qquad(s\le t<\sigma).
\]
Every retained coefficient remains less than $B/4$ because all update
indices in the block satisfy \eqref{eq:smallcondition}.

It remains to verify the rank reduction.
Let $V=\spn\{x_1,\ldots,x_n\}$, so that $\dim V=r$.
Equation \eqref{eq:positivecombination} implies $u\in V$.
The kernel of $P$ on $V$ is $\spn\{u\}$, and hence $\dim P(V)=r-1$.
Because $y_i=Px_i\in P(V)$ for $i\in G$,
\[
  1\le r_G:=\dim\spn\{y_i:i\in G\}\le r-1.
\]

After shifting time by $s$ and relabeling the retained samples, all
hypotheses of the auxiliary estimate \eqref{up:auxbound} now hold with
\[
  m=|G|,
  \qquad \mu=\frac\gamma2,
  \qquad c=\frac3{64},
  \qquad K=\frac{4n}{\gamma},
  \qquad E=1.
\]
The auxiliary estimate gives
\begin{equation}\label{eq:blocklength}
  \sigma-s
  \le C_{n,\gamma}(1+\log B)^{r_G-1}
  \le C_{n,\gamma}(1+\log B)^{r-2}.
\end{equation}
To make the constants uniform, take the maximum of the auxiliary estimate's
constants and thresholds over $1\le m\le n-1$ with the displayed scalar
parameters. This maximum depends only on $n$ and $\gamma$.
The recursive sequence starts at $z_s=0$ because we recomputed the
projection at this particular block start.
The endpoint $\sigma$ need not satisfy a coefficient or norm bound for the
application of the auxiliary estimate. Thus the last update of the block is
included even if it ends at a large coefficient or at the transition time.

\proofstep{Step 7. Summing the interval bounds.}
The update indices $0,\ldots,\tau_\eta-1$ form a disjoint union of the
$N_{\mathrm{large}}$ indices counted in \eqref{eq:largecount} and the
blocks constructed above. Consequently,
\begin{align*}
  \tau_\eta
  &=N_{\mathrm{large}}+\sum_{\text{blocks }[s,\sigma)}(\sigma-s)\\
  &\le\frac{8n}{\gamma^2}
    +\left(\frac{8n}{\gamma^2}+1+\frac{16}{\gamma^4}\right)
       C_{n,\gamma}(1+\log B)^{r-2}\\
  &\le C_{n,\gamma}(1+\log B)^{r-2}.
\end{align*}
The constant in the last line has been enlarged. This is allowed because
$r\ge2$ and $B\ge\e$, so the logarithmic factor is at least one.
Combining this with Step 3 gives
\[
  \tau_\eta\le
  C_{n,\gamma}(1+\log B)^{\max\{r-2,0\}}.
\]
Since $B=\eta/n$ and $n\ge2$,
\[
  1+\log B\le1+\log\eta.
\]
Because $1\le r\le\min\{n,d\}$ and $n,d\ge2$, we also have
\[
 \max\{r-2,0\}\le\min\{n-2,d-2\}.
\]
This proves the claimed upper bound for $\tau_\eta(D)$. Together with Lemma~\ref{lem:sketch-stability}, it also gives the upper bound for $\sigma_\eta(D)$.

All thresholds imposed directly on the original trajectory depend only on
$n$ and $\gamma$. In the auxiliary estimate, the recursive sample count
decreases at every application, and all constants depend only on its
listed scalar parameters. The outer application takes a maximum over
only $n-1$ possible sample counts. Thus one value of $\eta_{n,\gamma}$
and one constant $C_{n,\gamma}$ work for every ambient dimension and every
dataset satisfying the given norm and margin conditions.

\section{Lower bound}
\label{app:lower}

We prove the lower bound in Theorem~\ref{thm:main}. Fix
\[
k=\min\{n-2,d-2\}.
\]
When $k\ge 1$, our goal is to construct a linearly separable dataset for which a strict loss increase occurs after order $(\log \eta)^k$ iterations. A strict increase at such a late time is stronger than merely delaying entry into a low loss set. It directly implies that permanent monotonicity has not started earlier.

The proof has two parts. We first construct a long exact trajectory for an auxiliary logistic system with $k+1$ interacting coordinates. The construction is recursive. A new coordinate stays exponentially small while the old dynamics evolve, but its logarithmic depth changes by a fixed fraction across each old episode. This lets one concatenate order $\log B$ copies of the previous construction, where
\[
B=\frac{\eta}{n}.
\]
At the last level, the new coordinate produces a large pulse near the end of the trajectory. We then realize the auxiliary trajectory, up to an exponentially small error, as the true gradient descent trajectory after the first update from $w_0=0$.

Throughout this section, $k$ is fixed. Constants may depend on $k$ and
on matrices or compact sets fixed at earlier stages of the construction.
They do not depend on $B$ or on the scale of the current episode.
All asymptotic statements are uniform over points in the stated compact
sets. The index $r$ below counts recursive levels. It is distinct from
the sample rank used in \Cref{app:upper}.

\subsection{Exact coefficient dynamics}
\label{sub:lowerdynamics}

Consider vectors $v_1,\ldots,v_m$ with Gram matrix
\[
K_{ij}=\ip{v_i}{v_j}.
\]
For a parameter vector $w$, write
\[
m_i=\ip{w}{v_i}
\]
for the margin of the $i$th vector. Define
\[
p_i=B\e^{-m_i},
\qquad
h_B(x)=\frac{x}{1+x/B}.
\]
The quantity $p_i$ is not the coefficient in the gradient update. The actual coefficient is
\[
\frac{B}{1+\e^{m_i}}
=
\frac{B\e^{-m_i}}{1+\e^{-m_i}}
=
h_B(p_i).
\]
Therefore one gradient update gives
\[
w^+=w+\sum_{i=1}^m h_B(p_i)v_i.
\]
Taking inner products with the vectors $v_i$ gives
\[
m^+=m+Kh_B(p),
\]
where $h_B$ acts coordinatewise. Since $p_i=B\e^{-m_i}$, we obtain the exact coefficient map
\begin{equation}
 p^+=p\odot \exp\bigl(-Kh_B(p)\bigr).
 \label{eq:coeffmap}
\end{equation}
Here $\odot$ and the exponential are coordinatewise.
We call a finite sequence satisfying \eqref{eq:coeffmap} an exact orbit.
For such a coefficient state, define the scaled loss by
\begin{equation}\label{eq:scaledloss}
L_B(p)=B\sum_i\log(1+p_i/B).
\end{equation}
When these vectors are samples in the original dataset, $L_B(p)$ equals
$\eta$ times their contribution to $F(w)$. Since $\eta>0$, comparing
scaled losses preserves the direction of a loss change.

It is convenient to use logarithmic coordinates
\[
s=\log p.
\]
Then the map becomes
\begin{equation}
 \Phi_{B,K}(s)=s-Kh_B(\e^s).
 \label{eq:logmap}
\end{equation}
Every auxiliary orbit constructed below satisfies \eqref{eq:coeffmap} exactly. We never replace logistic loss by exponential loss.

We will repeatedly use
\begin{equation}
0\le x-h_B(x)=\frac{x^2}{B+x}\le \frac{x^2}{B},
\qquad
h_B'(x)=\frac{1}{(1+x/B)^2}.
\label{eq:hbasic}
\end{equation}
If
\[
t_B(x)=xh_B'(x)=\frac{x}{(1+x/B)^2},
\]
then a direct derivative calculation gives
\begin{equation}
\left|\frac{d}{du}\log t_B(\e^u)\right|
=
\left|\frac{B-\e^u}{B+\e^u}\right|
\le 1.
\label{eq:tloglip}
\end{equation}
This simple bound is what makes the local inverse estimate below uniform even when some coordinates are exponentially small.

\subsection{A uniform local inverse}
\label{sub:lowerinverse}

The recursive construction is most naturally built backward. We therefore need a local inverse whose size does not shrink when one coordinate becomes extremely small.

\begin{lemma}
\label{lem:appendixinverse}
Fix $0<\kappa<H<\infty$. There are constants $\delta>0$, $R>1$, $\alpha>0$, and $L\ge 2$, depending only on $\kappa$ and $H$, with the following property. Suppose
\[
\kappa I\preceq K\preceq HI
\]
and every coordinate of $p$ belongs to
\[
(0,\delta]\cup [R,B].
\]
Then $\Phi_{B,K}$ has a local inverse on the ball of radius $\alpha$ around $\Phi_{B,K}(\log p)$. This inverse sends the center to $\log p$ and is $L$ Lipschitz.
\end{lemma}

\begin{proof}
Set
\[
s_0=\log p,
\qquad
D=\diag\bigl(t_B(p_1),\ldots,t_B(p_m)\bigr).
\]
Differentiating \eqref{eq:logmap} gives
\[
J:=D\Phi_{B,K}(s_0)=I-KD.
\]
We first prove that $J^{-1}$ has a norm bound depending only on $\kappa$ and $H$.

Let
\[
A=\{i:p_i\ge R\},
\qquad
S=\{1,\ldots,m\}\setminus A.
\]
For $i\in A$, the assumptions $R\le p_i\le B$ imply
\[
t_B(p_i)=\frac{p_i}{(1+p_i/B)^2}\ge \frac{p_i}{4}\ge \frac{R}{4}.
\]
Hence
\[
D_A\succeq \frac{R}{4}I.
\]
For $i\in S$, we have $p_i\le\delta$, so
\[
0\preceq D_S\preceq \delta I.
\]
Assume first that both $A$ and $S$ are nonempty. In block form,
\[
J=
\begin{pmatrix}
E & -K_{AS}D_S\\
-K_{SA}D_A & I-K_{SS}D_S
\end{pmatrix},
\qquad
E=I-K_{AA}D_A.
\]
Because every principal submatrix of $K$ satisfies the same lower spectral bound,
\[
K_{AA}\succeq \kappa I.
\]
Choose $R\ge 8/\kappa$. Since $D_A^{-1}\preceq 4I/R$, we obtain
\[
K_{AA}-D_A^{-1}\succeq \frac{\kappa}{2}I.
\]
Using
\[
E=(D_A^{-1}-K_{AA})D_A,
\]
we get
\[
E^{-1}=D_A^{-1}(D_A^{-1}-K_{AA})^{-1}.
\]
Therefore
\begin{equation}
\norm{E^{-1}}\le \frac{8}{\kappa R},
\qquad
\norm{D_AE^{-1}}\le \frac{2}{\kappa}.
\label{eq:Ebound}
\end{equation}

Now solve $Jz=y$. The first block equation gives
\[
z_A=E^{-1}y_A+E^{-1}K_{AS}D_Sz_S.
\]
Substituting this expression into the second block equation gives
\[
S_0z_S=y_S+K_{SA}D_AE^{-1}y_A,
\]
where
\[
S_0
=
I-
\bigl(K_{SS}+K_{SA}D_AE^{-1}K_{AS}\bigr)D_S.
\]
Every submatrix of $K$ has operator norm at most $H$. Using \eqref{eq:Ebound},
\[
\norm{I-S_0}
\le
\left(H+\frac{2H^2}{\kappa}\right)\delta.
\]
Choose $\delta$ small enough that the right hand side is at most $1/2$. Then the geometric series for $S_0^{-1}$ converges and
\[
\norm{S_0^{-1}}\le 2.
\]
For clarity, the two block equations give the explicit bounds
\[
\norm{z_S}\le2\left(1+\frac{2H}{\kappa}\right)\norm y,
\qquad
\norm{z_A}\le\frac{8}{\kappa R}
\left(\norm y+H\delta\norm{z_S}\right).
\]
Since $R\ge8/\kappa$ and we may take $\delta\le1$, summing these
bounds shows that
\[
\norm{J^{-1}}\le M
\]
for a constant $M$ that depends only on $\kappa$ and $H$.

If $A$ is empty, then $\norm{KD}\le H\delta\le 1/2$ after decreasing $\delta$, so $I-KD$ is inverted by a geometric series. If $S$ is empty, the estimate follows directly from \eqref{eq:Ebound}. Thus the same bound on $J^{-1}$ holds in all cases.

We next show that the inverse exists on a ball whose radius is
independent of the sizes of the small coordinates. If $\norm{v}\le\rho$,
then $|v_i|\le\rho$ for every coordinate. By \eqref{eq:tloglip},
\[
\e^{-\rho}
\le
\frac{t_B(\e^{s_{0,i}+v_i})}{t_B(\e^{s_{0,i}})}
\le
\e^{\rho}.
\]
Thus the derivative at the displaced point is
\[
D\Phi_{B,K}(s_0+v)=I-KD(I+Q_v),
\qquad \norm{Q_v}\le\e^\rho-1,
\]
where $Q_v$ is diagonal. Since $J=I-KD$, we have
\[
J^{-1}KD=J^{-1}-I.
\]
Therefore
\[
\begin{aligned}
J^{-1}\bigl(D\Phi_{B,K}(s_0+v)-J\bigr)
&=-J^{-1}KDQ_v\\
&=-(J^{-1}-I)Q_v.
\end{aligned}
\]
Choose $0<\rho\le 1$ so that
\[
(M+1)(\e^\rho-1)\le \frac14.
\]
Then for every $s$ in the closed ball $\norm{s-s_0}\le\rho$,
\begin{equation}
\norm{I-J^{-1}D\Phi_{B,K}(s)}\le \frac14.
\label{eq:contractionderivative}
\end{equation}

Take a target $y$ with
\[
\norm{y-\Phi_{B,K}(s_0)}\le \frac{\rho}{4M}
\]
and define
\[
\Psi_y(s)=s+J^{-1}\bigl(y-\Phi_{B,K}(s)\bigr).
\]
At the center,
\[
\norm{\Psi_y(s_0)-s_0}
\le
M\norm{y-\Phi_{B,K}(s_0)}
\le
\frac{\rho}{4}.
\]
By \eqref{eq:contractionderivative}, $\Psi_y$ has Lipschitz constant at most $1/4$ on the ball. Hence for $\norm{s-s_0}\le\rho$,
\[
\norm{\Psi_y(s)-s_0}
\le
\frac{\rho}{4}+\frac14\norm{s-s_0}
\le
\frac{\rho}{2}.
\]
Thus $\Psi_y$ maps the closed ball into itself and is a contraction.
Starting at $s_0$ and repeatedly applying $\Psi_y$, the distances between
successive iterates decrease by a factor at most $1/4$. Their sum is
finite, so the iterates converge in the closed ball. Continuity gives a
fixed point, and the same contraction inequality makes it unique.
Its defining identity is exactly
\[
\Phi_{B,K}(s)=y.
\]

If $y'$ is another target and $s'$ is its fixed point, then
\[
\norm{s-s'}
\le
\frac14\norm{s-s'}+M\norm{y-y'}.
\]
Therefore
\[
\norm{s-s'}\le 2M\norm{y-y'}.
\]
Taking
\[
\alpha=\frac{\rho}{4M},
\qquad
L=\max\{2,2M\}
\]
proves the claim. The inverse is continuously differentiable on the
interior of its target ball. Indeed, \eqref{eq:contractionderivative}
implies that $D\Phi_{B,K}$ is invertible throughout the construction,
so the inverse function theorem applies at each selected solution.
Differentiating $\Phi_{B,K}(\mathcal R(y))=y$ gives
$D\mathcal R(y)=D\Phi_{B,K}(\mathcal R(y))^{-1}$.
\end{proof}

\subsection{Adding one negligible coordinate}
\label{sub:lowerextension}

The next lemma says that a coordinate can be added without changing a previously constructed orbit by more than an exponentially small amount. This is the main device that allows us to build the hierarchy one coordinate at a time.

\begin{lemma}
\label{lem:appendixextension}
Fix $C\ge1$, $q\ge0$, and $a_0>0$. Let $B\ge1$ and $T\ge1$.
Let
\[
\widetilde K=
\begin{pmatrix}
K & -b\\
-b^\top & 1
\end{pmatrix},
\qquad b\ge 0,
\]
and suppose that both $K$ and $\widetilde K$ satisfy common fixed positive
upper and lower spectral bounds. Let $p_0,\ldots,p_T$ be an exact orbit
for $K$. Assume that every coordinate of every $p_t$ lies in the ranges
of Lemma~\ref{lem:appendixinverse} for these common bounds and that, for
a scale $D\ge1$,
\[
T\le C(1+\log D)^q,
\qquad
\max_{t,i}p_{i,t}\le CD,
\qquad
\sum_{t<T}\sum_i h_B(p_{i,t})\le CD.
\]
Let $Y>0$ be the prescribed terminal value of the new coordinate. Assume either
\[
\log Y\le -a_0D
\]
or
\[
|\log Y|\le C\log D,
\qquad
b^\top h_B(p_{T-1})\ge a_0D,
\]
Then, for all sufficiently large $D$, there is an exact orbit $(\widehat p_t,g_t)_{t=0}^T$ for $\widetilde K$ ending at $(p_T,Y)$ such that
\[
\max_{0\le t<T}
\left(\norm{\log\widehat p_t-\log p_t}+g_t\right)
\le \exp(-cD)
\]
for a constant $c>0$ depending only on the fixed parameters and spectral
bounds.
\end{lemma}

\begin{proof}
For reference, the first terminal condition is
\begin{equation}
\log Y\le -a_0D.
\label{eq:terminalsmall}
\end{equation}
and the second is
\begin{equation}
|\log Y|\le C\log D,
\qquad
b^\top h_B(p_{T-1})\ge a_0D.
\label{eq:terminalpulse}
\end{equation}
We construct an approximate sequence first and then correct it backward
using Lemma~\ref{lem:appendixinverse}.

We first keep the old orbit exactly equal to $p_t$ and solve only for an approximate new coordinate. Set $\bar g_T=Y$. For $t=T-1,\ldots,0$, choose the small solution of
\begin{equation}
\log \bar g_t-h_B(\bar g_t)
=
\log \bar g_{t+1}-b^\top h_B(p_t).
\label{eq:gatebackward}
\end{equation}
We need to verify that this small solution exists at every step.

For a sufficiently negative number $v$, consider
\[
\log x-h_B(x)=v.
\]
On $0<x\le 1/4$ the derivative with respect to $\log x$ is
\[
1-t_B(x)\ge \frac34.
\]
At $x=\e^v$, the left hand side is at most $v$. At $x=\exp(v+2\e^v)$, the left hand side is at least $v$ for all sufficiently negative $v$. Therefore there is a unique solution in $(0,1/4)$ and it satisfies
\begin{equation}
v\le \log x\le v+2\e^v.
\label{eq:smallscalarinverse}
\end{equation}

Under \eqref{eq:terminalpulse}, the right hand side of \eqref{eq:gatebackward} at $t=T-1$ is at most
\[
C\log D-a_0D\le -\frac{a_0D}{2}
\]
for large $D$. Under \eqref{eq:terminalsmall}, it is at most $-a_0D$ because $b\ge0$. Thus \eqref{eq:smallscalarinverse} gives
\[
\bar g_{T-1}\le \exp(-a_0D/3)
\]
when $D$ is large enough.

At an earlier step, the nonnegativity of $b$ gives
\[
\log \bar g_t
\le
\log \bar g_{t+1}+2\bar g_{t+1}.
\]
Let
\[
q_D=\exp(-a_0D/3).
\]
Suppose the coordinates already constructed backward are at most $2q_D$. Then the total possible increase in their logarithms over all remaining steps is at most
\[
4Tq_D.
\]
Since $T$ is bounded by a fixed power of $\log D$, we have $4Tq_D<\log 2$ for all sufficiently large $D$. Starting from $\bar g_{T-1}\le q_D$, backward induction yields
\begin{equation}
\bar g_t\le 2q_D
\qquad
0\le t<T.
\label{eq:approxgsmall}
\end{equation}
This also confirms that every scalar inverse was taken on the intended small branch.

The sequence $(p_t,\bar g_t)$ is not yet an exact orbit of the enlarged system because the new coordinate exerts a small force on the old coordinates. In logarithmic variables let
\[
z_t=(\log p_t,\log \bar g_t).
\]
Equation \eqref{eq:gatebackward} makes the last coordinate equation exact. The old coordinates satisfy
\[
\Phi_{B,\widetilde K}(z_t)
=
z_{t+1}+\bigl(bh_B(\bar g_t),0\bigr).
\]
By \eqref{eq:approxgsmall}, the defect has norm at most $Cq_D$.

Start from the exact terminal point $\widehat z_T=z_T$. At each
$t<T$, the old coordinates $p_t$ satisfy the required ranges and
$\bar g_t\le2q_D\le\delta$. Thus Lemma~\ref{lem:appendixinverse}
applies at $z_t$ for the enlarged matrix. No inverse is needed at the
terminal state, so $Y$ need not be in either coordinate range.
Move backward using these local inverses, and let
\[
\varepsilon_t=\norm{\widehat z_t-z_t}.
\]
If $L$ is the uniform Lipschitz constant of the local inverse, then
\[
\varepsilon_t
\le
L\varepsilon_{t+1}+CLq_D.
\]
Since $\varepsilon_T=0$, iteration gives
\[
\varepsilon_t
\le
CT L^Tq_D.
\]
Taking logarithms and using $T\le C(1+\log D)^q$ gives
\begin{align*}
\log(CT L^Tq_D)
&=-\frac{a_0D}{3}+\log C+\log T+T\log L\\
&\le-\frac{a_0D}{3}+O((1+\log D)^q)\\
&\le-\frac{a_0D}{6}
\end{align*}
for sufficiently large $D$. The last inequality follows from
$(1+\log D)^q=o(D)$ for fixed $q$. Thus, with $c=a_0/6$,
\begin{equation}
\max_{0\le t\le T}\varepsilon_t
\le
\exp(-cD).
\label{eq:extensionerror}
\end{equation}
To justify every inverse application, choose $D$ so large that
$CT L^Tq_D+Cq_D<\alpha$, where $\alpha$ is the target radius in
Lemma~\ref{lem:appendixinverse}. If the correction has been constructed
through time $t+1$, the next target is at distance at most
$\varepsilon_{t+1}+Cq_D<\alpha$ from
$\Phi_{B,\widetilde K}(z_t)$. The inverse therefore exists at time $t$
and gives the stated recurrence. This backward induction constructs the
whole orbit. Since the logarithm of the corrected new coordinate differs
from $\log\bar g_t$ by at most $\exp(-cD)$, after decreasing $c$,
\[
\max_{t<T}g_t\le \exp(-cD).
\]

Write the exact new coordinate as $g_t$. Summing its logarithmic update gives
\[
\log g_T-\log g_0
=
\sum_{t<T}b^\top h_B(\widehat p_t)
-
\sum_{t<T}h_B(g_t).
\]
Using $g_T=Y$,
\[
-\log g_0
=
\sum_{t<T}b^\top h_B(\widehat p_t)
-
\log Y
-
\sum_{t<T}h_B(g_t).
\]
For $|v|\le\epsilon\le1$, the derivative bound
$t_B(x)\le h_B(x)$ gives
\[
|h_B(x\e^v)-h_B(x)|\le(\e^\epsilon-1)h_B(x)
\le2\epsilon h_B(x).
\]
Apply this with the logarithmic error in \eqref{eq:extensionerror}.
The old total coefficient mass is at most $CD$, so its change is at most
$CD\exp(-cD)$. The new coordinate contributes at most
$T\exp(-cD)$ before time $T$. Decreasing $c$ absorbs both factors.
Therefore
\begin{equation}
-\log g_0
=
b^\top\sum_{t<T}h_B(p_t)
-
\log Y
+
O(\exp(-cD)).
\label{eq:initialgatedepth}
\end{equation}
This identity will determine the scale of the new gate in the recursive construction.
\end{proof}

\subsection{An auxiliary orbit with a late loss increase}
\label{sub:lowerorbit}

We now construct the orbit underlying Lemmas~\ref{lem:sketch-base} and
\ref{lem:sketch-hierarchy} and Proposition~\ref{prop:sketch-late} in
\Cref{sec:proof-sketch}. An orbit segment at one recursive level
will be called an episode. Its endpoints and length will be specified
in the proof. The loss comparisons for this auxiliary trajectory use
$L_B$ from \eqref{eq:scaledloss}. The corresponding contribution to the
original empirical loss is $L_B/\eta$.

\begin{proposition}
\label{prop:appendixorbit}
For each integer $k\ge1$, there are a positive definite matrix $K$ of
size $k+1$ with unit diagonal and a compact set
$\mathcal P\subset(0,\infty)^{k+1}$ such that, for every sufficiently
large $B$, an exact orbit starting in $\mathcal P$ satisfies
\[
L_B(p_{t+1})>L_B(p_t)
\quad\text{at some }t=\Theta((\log B)^k).
\]
\end{proposition}

\begin{proof}
The proof is recursive. We begin with two coordinates and then add one
coordinate at a time. Following \Cref{sec:proof-sketch}, we call
each added coordinate a gate. When its value is $\e^{-E}$, its depth is
$E$. We will also prove that the constructed orbit has length
$T=\Theta((\log B)^k)$ and total coefficient mass $O(B^{1/4})$.
These estimates will be used to realize the orbit from $w_0=0$ in
\Cref{sub:lowerrealization}.

\paragraph{Step 1. The two coordinate orbit.}
Fix
\[
a=\frac45,
\qquad
K_1=
\begin{pmatrix}
1&-a\\
-a&1
\end{pmatrix}.
\]
The negative off diagonal entries mean that a large coefficient in one coordinate suppresses itself and increases the other coordinate. We now make this precise by constructing predecessors.

Let $(X,Y)$ be a target with $X$ large, $X^2\le B$, and
\begin{equation}
\left|
\log Y-\log(X/a)+ah_B(X)
\right|
\le H_0,
\label{eq:pairtarget}
\end{equation}
where $H_0$ is fixed. We seek a predecessor $(u,v)$. The inverse equations from \eqref{eq:coeffmap} are
\begin{align}
\log u-h_B(u)&=\log X-ah_B(v),
\label{eq:pairinv1}\\
\log v-h_B(v)+ah_B(u)&=\log Y.
\label{eq:pairinv2}
\end{align}
We first solve for $u$ as a function of $v$ on
$[aX/2,3aX/2]$. On this interval, $v$ is of order $X$, and
$v\le B$ for large $X$ because $X^2\le B$. Therefore
\[
h_B(v)\ge \frac{v}{2}.
\]
The right hand side of \eqref{eq:pairinv1} is then at most $-cX$. By the scalar inverse estimate \eqref{eq:smallscalarinverse}, there is a unique small solution and
\begin{equation}
u(v)\le \exp(-cX).
\label{eq:usmall}
\end{equation}
Implicit differentiation of \eqref{eq:pairinv1} gives
\[
u'(v)
=
-\frac{ah_B'(v)}{1/u(v)-h_B'(u(v))}<0.
\]

Substitute $u(v)$ into \eqref{eq:pairinv2} and define
\[
\Psi(v)=\log v-h_B(v)+ah_B(u(v))-\log Y.
\]
For $v$ of order $X$,
\[
h_B'(v)=1+O(1/X)
\]
because $v/B=O(1/X)$. Hence
\[
\Psi'(v)
=
\frac1v-h_B'(v)+ah_B'(u(v))u'(v)
\le
-\frac12
\]
for all sufficiently large $X$.

At $v=aX$, use \eqref{eq:pairtarget}. The residual is
\[
\Psi(aX)
=
2\log a+ah_B(X)-h_B(aX)
-
\Bigl(\log Y-\log(X/a)+ah_B(X)\Bigr)
+
ah_B(u(aX)).
\]
The quantity in parentheses is bounded by $H_0$. Also, by \eqref{eq:hbasic},
\[
|ah_B(X)-h_B(aX)|
\le
\frac{aX^2}{B}+\frac{a^2X^2}{B}
=O(1).
\]
Finally $h_B(u(aX))$ is exponentially small. Thus $|\Psi(aX)|\le C$
for a fixed $C>0$. Set
\[
v_-=aX-(2C+2),
\qquad v_+=aX+(2C+2).
\]
For sufficiently large $X$, both endpoints lie in $[aX/2,3aX/2]$,
where the derivative bound above holds. Integrating that bound gives
\begin{align*}
\Psi(v_-)
&\ge\Psi(aX)+\frac{aX-v_-}{2}\ge-C+(C+1)=1,\\
\Psi(v_+)
&\le\Psi(aX)-\frac{v_+-aX}{2}\le C-(C+1)=-1.
\end{align*}
The intermediate value theorem therefore gives a root in $(v_-,v_+)$.
Since $\Psi'(v)<0$, it is the unique root in this interval and satisfies
\begin{equation}
v=aX+O(1).
\label{eq:vscale}
\end{equation}
Together with \eqref{eq:usmall}, this gives the desired predecessor.

There is an exact expression for the error in the corresponding relation for the new pair. Rearranging \eqref{eq:pairinv1},
\[
\log u=\log X-ah_B(v)+h_B(u).
\]
Therefore
\[
\log u-\log(v/a)+ah_B(v)
=
\log\frac{aX}{v}+h_B(u).
\]
By \eqref{eq:vscale}, the first term on the right is $O(1/X)$, and the second is exponentially small. Hence the new pair satisfies the same type of condition as \eqref{eq:pairtarget}, with an error that is in fact smaller.

We may now repeat the selected inverse. At every step the dominant coordinate switches from one component to the other. If $X_j$ denotes the dominant value after the $j$th backward step, then
\begin{equation}
X_{j+1}=aX_j+O(1).
\label{eq:dominantrec}
\end{equation}
Choose a large fixed stopping level $L_0$. Stop when the dominant value first becomes at most $2L_0$. Iterating \eqref{eq:dominantrec} gives
\[
X_j=a^jX+O\left(\sum_{i=0}^{j-1}a^i\right)=a^jX+O(1).
\]
The first crossing has $3L_0/2\le X_J\le2L_0$ for large enough $L_0$,
because $X_{J-1}>2L_0$ and $X_J=aX_{J-1}+O(1)$ with $a=4/5$.
Taking $L_0$ larger than the fixed error in the last display shows that
$a^JX$ is between two positive fixed constants. Therefore
\begin{equation}
J=\frac{\log X}{|\log a|}+O(1).
\label{eq:pairlength}
\end{equation}
Reversing the backward sequence produces an exact forward orbit of
length $\Theta(\log X)$, proving Lemma~\ref{lem:sketch-base}.

We also record two estimates that will be needed later. Summing the dominant values in backward time gives
\[
\sum_{j=1}^J X_j
=
\frac{a}{1-a}X+O(J).
\]
The small coordinates are uniformly bounded, so they contribute $O(J)$.
Also $X_j\le C a^jX$ before the stopping index, after increasing $C$.
Replacing $X_j$ by $h_B(X_j)$ therefore changes the sum by at most
\[
\frac1B\sum_{j=1}^JX_j^2=O(1)
\]
because $X^2\le B$. Consequently the total coefficient mass of the exact pair orbit is
\begin{equation}
\sum_{t<T}\bigl(h_B(p_{1,t})+h_B(p_{2,t})\bigr)
=
\frac{a}{1-a}X+O(\log X).
\label{eq:pairmass}
\end{equation}
The initial point obtained at the fixed stopping scale lies in a fixed compact subset of $(0,\infty)^2$.

When $k\ge2$, we also prepare the finite collection of pair targets that will be needed by the later recursion. For
\[
0\le j\le k-2,
\]
let $Q_2^j$ be the union of the following set and the set obtained by swapping its two coordinates:
\begin{equation}
\left\{
(X,Y):
 a^{2j}L_0\le X\le 3a^{-2j}L_0,
\ \left|\log Y-\log(X/a)+aX\right|\le 2+j
\right\}.
\label{eq:paircompact}
\end{equation}
Each $Q_2^j$ is a compact subset of the positive quadrant with nonempty interior. The difference between $h_B(X)$ and $X$ tends to zero uniformly on these fixed compact sets as $B$ tends to infinity. Hence the predecessor estimate \eqref{eq:vscale} and the residual estimate following it also hold uniformly over every $Q_2^j$. If $L_0$ is chosen large enough, the selected predecessor of every point in $Q_2^j$ lies a fixed positive distance inside $Q_2^{j+1}$ whenever $j<k-2$. The reason is simple. Its dominant coordinate is $aX+O(1)$, which lies strictly inside the enlarged interval at the next index, and its new residual has magnitude less than one. Since there are only finitely many compact sets, Lemma~\ref{lem:appendixinverse} gives one positive target radius and one derivative bound for every inverse branch that will be used later.

Taking $X=B^{1/4}$ already proves the proposition when $k=1$, except for the strict loss increase. We verify that at the end of the proof.

\paragraph{Step 2. The first gate.}
Assume now that $k\ge2$. We add a third coordinate. Fix
\[
c=\frac15,
\qquad
K_2=
\begin{pmatrix}
1&-a&-c\\
-a&1&-c\\
-c&-c&1
\end{pmatrix}.
\]
The eigenvalues are
\[
1+a,
\qquad
\frac{3+\sqrt6}{5},
\qquad
\frac{3-\sqrt6}{5}.
\]
In particular, $K_2$ is positive definite with a fixed spectral gap from zero.

We call the third coordinate the gate. Write the terminal gate as $\exp(-E)$. The quantity $E$ is its logarithmic depth. We now show that one exact three coordinate episode changes this depth by a fixed fraction while taking $\Theta(\log E)$ updates.

Fix
\[
(P,Q)\in Q_2^j
\]
for some $0\le j\le k-2$. All estimates below are uniform over these finitely many compact target sets. Assume
\[
E_0\le E\le B^{1/4}.
\]
We first solve the last update backward. Seek a predecessor of the form
\[
(M,\e^{-\ell},R).
\]
The exact equations are
\begin{align}
\log P
&=
\log M-h_B(M)+ah_B(\e^{-\ell})+ch_B(R),
\label{eq:firstgate1}\\
\log Q
&=
-\ell+ah_B(M)-h_B(\e^{-\ell})+ch_B(R),
\label{eq:firstgate2}\\
-E
&=
\log R+ch_B(M)+ch_B(\e^{-\ell})-h_B(R).
\label{eq:firstgate3}
\end{align}
Ignoring logarithmic terms and the exponentially small coordinate suggests
\[
M\approx cR,
\qquad
R-cM\approx E.
\]
Solving these leading equations gives
\[
\bar M=\frac{cE}{1-c^2},
\qquad
\bar R=\frac{E}{1-c^2}.
\]
Equation \eqref{eq:firstgate2} then suggests
\[
\bar\ell=(1+a)\bar M.
\]
Let $f(M,R,\ell)$ be the vector obtained by moving the left hand sides of \eqref{eq:firstgate1} through \eqref{eq:firstgate3} to the right. At the center $(\bar M,\bar R,\bar\ell)$, the terms of order $E$ cancel by construction. For example,
\[
\bar M-c\bar R=0,
\qquad
\bar R-c\bar M=E.
\]
The remaining terms contain $\log \bar M$, $\log \bar R$, the bounded numbers $\log P$ and $\log Q$, and the logistic corrections from \eqref{eq:hbasic}. Since $\bar M,\bar R=\Theta(E)$ and $E\le B^{1/4}$,
\[
\frac{\bar M^2}{B}+\frac{\bar R^2}{B}=O(1).
\]
The terms containing $\e^{-\bar\ell}$ are exponentially small. Hence
\[
\norm{f(\bar M,\bar R,\bar\ell)}\le C\log E.
\]

We next compute the derivative. The three rows are
\[
\left(
\frac1M-h_B'(M),
ch_B'(R),
-at_B(\e^{-\ell})
\right),
\]
\[
\left(
ah_B'(M),
ch_B'(R),
-1+t_B(\e^{-\ell})
\right),
\]
and
\[
\left(
ch_B'(M),
\frac1R-h_B'(R),
-ct_B(\e^{-\ell})
\right).
\]
On a ball of radius $C_1\log E$ around the center, $M$, $R$, and $\ell$ remain positive and of order $E$. By \eqref{eq:hbasic},
\[
h_B'(M)=1+O(E/B),
\qquad
h_B'(R)=1+O(E/B).
\]
Also $1/M+1/R=O(1/E)$ and $t_B(\e^{-\ell})$ is exponentially small. Therefore the derivative differs by
\[
O(E^{-1}+E/B+\exp(-c_0E))
\]
from
\[
J_0=
\begin{pmatrix}
-1&c&0\\
a&c&-1\\
c&-1&0
\end{pmatrix}.
\]
Its determinant is $1-c^2>0$.

Choose $C_1$ large enough that
\[
\norm{J_0^{-1}f(\bar M,\bar R,\bar\ell)}
\le
\frac{C_1\log E}{2}.
\]
Then increase the lower threshold on $E$ so that
\[
\norm{J_0^{-1}(Df-J_0)}\le \frac12
\]
throughout the ball of radius $C_1\log E$. The map
\[
x\mapsto x-J_0^{-1}f(x)
\]
then sends this ball into itself and has Lipschitz constant at most $1/2$. The contraction theorem gives an exact solution with
\begin{equation}
M=\frac{cE}{1-c^2}+O(\log E),
\qquad
R=\frac{E}{1-c^2}+O(\log E),
\qquad
\ell=(1+a)M+O(\log E).
\label{eq:firstgatelast}
\end{equation}

We solve one more update backward. Seek a predecessor
\[
(\e^{-A},N,\e^{-H}).
\]
The exact equations are
\begin{align}
\log M
&=
-A-h_B(\e^{-A})+ah_B(N)+ch_B(\e^{-H}),
\label{eq:firstgateprev1}\\
-\ell
&=
\log N+ah_B(\e^{-A})-h_B(N)+ch_B(\e^{-H}),
\label{eq:firstgateprev2}\\
\log R
&=
-H+ch_B(\e^{-A})+ch_B(N)-h_B(\e^{-H}).
\label{eq:firstgateprev3}
\end{align}
The leading solution has
\[
N\approx \ell,
\qquad
A\approx a\ell-\log M,
\qquad
H\approx c\ell-\log R.
\]
At this center the terms of order $E$ again cancel. The remaining logarithmic terms give a residual of size $O(\log E)$. If the variables are ordered as $(N,A,H)$, the derivative rows converge to
\[
(a,-1,0),
\qquad
(-1,0,0),
\qquad
(c,0,-1).
\]
Indeed, the terms containing $\e^{-A}$ and $\e^{-H}$ are exponentially small, while $h_B'(N)=1+O(E/B)$ and $1/N=O(1/E)$. Thus the limiting derivative is
\[
J_1=
\begin{pmatrix}
a&-1&0\\
-1&0&0\\
c&0&-1
\end{pmatrix},
\]
which has determinant $1$. On a ball of radius $C_2\log E$, the derivative error is
\[
O(E^{-1}+E/B+\exp(-c_0E)).
\]
Choose $C_2$ to dominate the central residual after multiplication by $J_1^{-1}$, and then increase the lower threshold on $E$ so that the derivative error gives contraction factor at most $1/2$. This gives an exact solution with
\begin{equation}
N=(1+a)M+O(\log E),
\qquad
A=aN-\log M+O(1),
\qquad
H=cN-\log R+O(1).
\label{eq:firstgateprev}
\end{equation}

We verify the target condition before applying the pair construction.
Equation~\eqref{eq:firstgateprev1} gives
\[
-A=\log M-ah_B(N)+h_B(\e^{-A})-ch_B(\e^{-H}).
\]
Thus the pair with dominant coordinate $N$ satisfies
\[
-A-\log(N/a)+ah_B(N)
=\log(aM/N)+h_B(\e^{-A})-ch_B(\e^{-H}).
\]
The right hand side tends to $\log(a/(1+a))$, whose absolute value is
less than one. Hence \eqref{eq:pairtarget} holds with a fixed residual
bound. Moreover, $N=\Theta(E)$ and $E\le B^{1/4}$ imply $N^2\le B$
for sufficiently large $B$. We may therefore apply the selected pair
inverse until its dominant coordinate returns to the fixed stopping
scale. By \eqref{eq:pairlength}, this prefix has length
$\Theta(\log E)$. By \eqref{eq:pairmass}, its coefficient mass is
\[
\frac{a}{1-a}N+O(\log E).
\]

The third coordinate at the end of this prefix is $\e^{-H}$.
We check all assumptions of Lemma~\ref{lem:appendixextension} at scale
$D=N$. The eigenvalues of $K_1$ and $K_2$ lie in $[1/16,2]$.
Choose $L_0$ large enough that every dominant pair coordinate is at
least the corresponding $R$ in Lemma~\ref{lem:appendixinverse}, and
every small coordinate is at most its $\delta$. The largest coordinate
is $O(N)$, which is less than $B$ for large $B$ because
$N=\Theta(E)$ and $E\le B^{1/4}$. The preceding pair estimates give
length $O(\log N)$ and total mass $O(N)$.
Finally, \eqref{eq:firstgateprev} gives $H\ge cN/2$, so the prescribed
terminal third coordinate satisfies \eqref{eq:terminalsmall}.
We may therefore apply the lemma with
\[
b=c(1,1)^\top.
\]
This produces an exact three coordinate prefix in which the first two coordinates differ from the pair orbit by only an exponentially small amount, and the third coordinate remains exponentially small throughout the prefix. The initial pair state of the unperturbed prefix lies a fixed positive distance inside $Q_2^0$. The correction tends to zero uniformly, so the corrected initial pair state remains in a fixed compact subset of the interior of $Q_2^0$.

Let $D$ be the gate depth at the beginning of the resulting three coordinate episode. Formula \eqref{eq:initialgatedepth} gives
\[
D
=
H+c\sum_{t<T}\bigl(h_B(p_{1,t})+h_B(p_{2,t})\bigr)+O(1).
\]
Substituting the pair mass estimate and \eqref{eq:firstgateprev},
\[
D
=
H+c\frac{a}{1-a}N+O(\log E)
=
\frac{c}{1-a}N+O(\log E).
\]
Using \eqref{eq:firstgatelast} and \eqref{eq:firstgateprev},
\[
\frac{D}{E}
=
\frac{c^2(1+a)}{(1-a)(1-c^2)}+O\left(\frac{\log E}{E}\right).
\]
For $a=4/5$ and $c=1/5$, the constant is $3/8$. Hence, for all sufficiently large $E$,
\begin{equation}
\frac{E}{4}\le D\le \frac{E}{2}.
\label{eq:firstgatedepthratio}
\end{equation}

We have proved more than the depth ratio. The episode has length $\Theta(\log E)$, its total coefficient mass is $O(E)$, and the last update has a gate coefficient of order $E$ because $R=\Theta(E)$. Before this last part, the gate is exponentially small. These estimates are exactly what will be needed in the recursive step.

\paragraph{Step 3. Adding further gates.}
For $3\le r\le k$, define
\begin{equation}
K_r=
\begin{pmatrix}
K_{r-1}&-\varepsilon_r e_r\\
-\varepsilon_r e_r^\top&1
\end{pmatrix},
\label{eq:Krrec}
\end{equation}
where $e_r$ is the last standard basis vector in $\R^r$. The new
coordinate is coupled only to the previous gate. We choose the couplings
recursively. Once the construction through level $r$ and its constant
$C_r$ have been fixed, choose $\varepsilon_{r+1}>0$ to satisfy both
bounds in \eqref{eq:epschoice} below. In particular,
$\varepsilon_{r+1}\le2^{-r-7}$, so for the fixed final level $k$,
\[
\sum_{j=3}^k\varepsilon_j
\le\sum_{r=2}^{k-1}2^{-r-7}
<\frac1{256}.
\]
When $k=2$, both sums are zero. Each coupling is fixed before constructing
the next level and does not depend on $B$ or the episode scale.
The off diagonal perturbation that turns $\operatorname{diag}(K_{r-1},1)$ into $K_r$ has operator norm $\varepsilon_r$. Therefore, for every unit vector $z$, its quadratic form changes by at most $\varepsilon_r$. Taking the minimum and maximum over unit vectors shows that the smallest and largest eigenvalues can move by at most $\varepsilon_r$ at this step. Since the smallest eigenvalue of $K_2$ is larger than $1/10$, the largest equals $9/5$, and the total later perturbation is less than $1/256$, we have
\begin{equation}
\frac1{16}I\preceq K_r\preceq 2I
\label{eq:spectraluniform}
\end{equation}
for every level that we use. This allows us to use the same type of local inverse estimate at every recursive step.

We now explain the induction. At level $r$, the matrix $K_r$ has $r+1$ coordinates. We write it as
\begin{equation}
K_r=
\begin{pmatrix}
K_{r-1}&-b_r\\
-b_r^\top&1
\end{pmatrix},
\qquad
b_2=c(1,1)^\top,
\qquad
b_r=\varepsilon_r e_r\quad(r\ge3).
\label{eq:Krblock}
\end{equation}
The last coordinate is the current gate. An endpoint has the form
\[
(q,\e^{-D}),
\]
where $q$ contains the first $r$ coordinates and stays in a fixed compact set.

There are two technical facts that must be preserved from one level to the next. First, the bounded profile must admit the few backward steps needed by later levels. Second, every coordinate must remain either uniformly small or uniformly large so that Lemma~\ref{lem:appendixinverse} applies. Since the final depth $k$ is fixed, only finitely many such backward steps are ever needed.

We implement the first fact with a short chain of compact sets
\[
Q_r^0\subset Q_r^1\subset\cdots\subset Q_r^{k-r}\subset(0,\infty)^r.
\]
For every $j<k-r$, each target in $Q_r^j$ has a selected predecessor for the bounded $r$ coordinate dynamics that lies in a fixed compact subset of the interior of $Q_r^{j+1}$. The selected inverse is defined on a logarithmic ball of a fixed positive radius, uniformly over the target. At level $r=2$, these compact sets are obtained directly from the pair construction. We enlarge the allowed interval for the dominant coordinate and the bounded residual in \eqref{eq:pairtarget} a fixed amount at each step. The estimates $v=aX+O(1)$ and
\[
\log u-\log(v/a)+ah_B(v)=O(1/X)
\]
show that the selected predecessor lies strictly inside the next enlarged set once the fixed stopping scale is chosen large enough. Compactness then gives one inverse radius that works uniformly on each set.

For the second fact, take $\delta$ and $R$ from
Lemma~\ref{lem:appendixinverse} using \eqref{eq:spectraluniform}.
Choose fixed numbers $\delta_*>0$ and $R_*>0$ such that
\[
\e^{k+2}\delta_*<\delta,
\qquad \e^{-k-2}R_*>R.
\]
Choose the pair stopping scale large enough that its dominant coordinate
is at least $R_*$ and its other coordinate is at most $\delta_*$,
including on all the compact pair sets. Choose every subsequent fixed
stopping depth large enough to give the same bounds for each newly
constructed coordinate. Each correction below has logarithmic size less
than one. Along any chain of constructions, a coordinate is corrected at
most once per added coordinate, so these strict inequalities allow all
corrections through level $k$. Concatenating episodes introduces no
additional correction at their common endpoints.

The upper endpoint $B$ requires a separate check. We maintain the bound
$p_{i,t}\le C_rE$ at level $r$, where $E\le B^{1/4}$. For the finitely
many levels, choose $B$ sufficiently large that
$\e^{k+2}\max_r C_r B^{1/4}<B$. Thus the corrected large coordinates
also remain below $B$. This proves that the inverse lemma's two ranges
are preserved whenever it is used.

The induction claim is as follows. For each $2\le r\le k$, there are
constants $\Lambda_r\ge4$, $c_r>0$, $C_r>0$, and $E_r^*>0$ that work
for every $0\le j\le k-r$ and every $q^+\in Q_r^j$. For all
$E_r^*\le E\le B^{1/4}$, there is an exact $K_r$ orbit from
\[
(q^-,\e^{-D})
\]
to
\[
(q^+,\e^{-E}).
\]
The initial bounded profile $q^-$ belongs to a fixed compact subset of
the interior of $Q_r^0$. Throughout each episode, including both
endpoints, every coordinate remains in the two ranges required by
Lemma~\ref{lem:appendixinverse}. The initial depth and the length satisfy
\begin{align}
\frac{E}{\Lambda_r}
&\le D\le \frac{E}{2},
\label{eq:inddepth}\\
c_r(\log E)^{r-1}
&\le T_r(E)\le C_r(\log E)^{r-1}.
\label{eq:indlength}
\end{align}
We also maintain estimates needed for the next level. After decreasing
$c_r$ and increasing $C_r$ if necessary, they are
\begin{equation}\label{eq:indmass}
\begin{gathered}
\sum_{t<T_r(E)}\sum_{i=1}^{r+1}h_B(p_{i,t})\le C_rE,
\qquad \max_{0\le t\le T_r(E),\,i}p_{i,t}\le C_rE,\\
h_B(p_{r+1,T_r(E)-1})\ge c_rE.
\end{gathered}
\end{equation}
The last inequality bounds the gate coefficient at the last update.
It also gives a lower bound for the gate's total mass, while the first
inequality gives its upper bound. Step 2 establishes these inequalities
for $r=2$: the two final states have coordinates $O(E)$, the earlier
pair mass is $O(E)$, and $R=\Theta(E)$ at the last update. Throughout
the proof, a coordinate is $p_{i,t}$ and its update coefficient is
$h_B(p_{i,t})$.

Assume the claim holds at level $r<k$. All constants and compact sets at level $r$ are now fixed. We construct level $r+1$.

\paragraph{Preparing the bounded target sets at the next level.}
With $C_r$ now fixed, choose the new coupling
\[
\varepsilon=\varepsilon_{r+1}>0
\]
small enough that
\begin{equation}
\varepsilon\le 2^{-r-7},
\qquad
\varepsilon\le \frac{1}{8\sqrt{C_r+1}}.
\label{eq:epschoice}
\end{equation}
The second inequality will later give the upper depth bound $D\le E/2$.

Fix $j\le k-r-1$ and a target $q^+\in Q_r^j$. For a large number $d$, consider the exact equation
\begin{equation}
\log q^+
=
\log q-K_{r-1}h_B(q)+b_rh_B(\e^{-d}).
\label{eq:qofdeq}
\end{equation}
Without the last term, this is precisely the bounded predecessor equation already available at level $r$. The perturbation has norm at most
\[
\norm{b_r}\e^{-d}.
\]
For large $d$, it lies inside the fixed inverse neighborhood. Hence \eqref{eq:qofdeq} has a selected solution, which we denote by $q(d)$. It lies in a fixed compact subset of the interior of $Q_r^{j+1}$.

We also need to know that $q(d)$ changes very little when $d$ changes. Let $\mathcal R$ denote the selected inverse in logarithmic coordinates. Then
\[
\log q(d)
=
\mathcal R\left(\log q^+-b_rh_B(\e^{-d})\right).
\]
Differentiate with respect to $d$. The derivative of $\mathcal R$ is uniformly bounded on the compact target set, and
\[
\left|\frac{d}{dd}h_B(\e^{-d})\right|
=t_B(\e^{-d})
\le \e^{-d}.
\]
Therefore
\begin{equation}
\norm{q'(d)}\le C\e^{-d}.
\label{eq:qderivative}
\end{equation}

Choose a constant $M\ge1$ that bounds
\[
b_r^\top h_B(q(d))
\]
on every branch that can occur. Such an $M$ exists because all $q(d)$ lie in finitely many fixed compact sets. Let
\[
H_0=M+3.
\]
Choose a fixed stopping depth $A>E_r^*$. Increase it so that
$A/2-kH_0-1$ is positive and is already large enough for
\eqref{eq:qderivative}, for $\e^{-d}$ to be in the small coordinate
range, and for $C\e^{-d}\le1/2$. This includes every depth in the
intervals used below.

For $0\le j\le k-r-1$, define the depth interval
\begin{equation}
I_j=
\left[
\frac{A}{2}-jH_0,
2\Lambda_rA+jH_0
\right]
\label{eq:depthinterval}
\end{equation}
and define
\[
Q_{r+1}^j
=
\left\{(q,\e^{-d}):q\in Q_r^j,\ d\in I_j\right\}.
\]
These are compact sets. We now verify the required bounded predecessor property for them before constructing any level $r+1$ episode.

Take a target $(q^+,\e^{-d^+})\in Q_{r+1}^j$ with $j<k-r-1$. A predecessor under the $K_r$ dynamics must have the form $(q(d),\e^{-d})$, where $q(d)$ solves \eqref{eq:qofdeq} and the last coordinate equation is
\begin{equation}
d
=
d^++b_r^\top h_B(q(d))-h_B(\e^{-d}).
\label{eq:boundeddepthfixedpoint}
\end{equation}
On the interval
\[
[d^+-1,d^++M+1],
\]
the right hand side of \eqref{eq:boundeddepthfixedpoint} lies between $d^+-1$ and $d^++M$. Its derivative has magnitude at most $C\e^{-d}$ by \eqref{eq:qderivative}. By the choice of $A$, this is at most $1/2$. Hence the right hand side is a contraction of the interval into itself and has a unique fixed point.

The interval $I_{j+1}$ is obtained from $I_j$ by enlarging both endpoints by $H_0=M+3$. Thus the new depth lies a fixed positive distance inside $I_{j+1}$. The point $q(d)$ lies a fixed positive distance inside $Q_r^{j+1}$. Therefore the full predecessor lies in a fixed compact subset of the interior of $Q_{r+1}^{j+1}$. Lemma~\ref{lem:appendixinverse} then gives a uniform inverse neighborhood. This proves the bounded predecessor property at level $r+1$.

\paragraph{Solving the last update of a new episode.}
Fix a terminal point
\[
(q^+,\e^{-d^+},\e^{-E})
\]
with $(q^+,\e^{-d^+})\in Q_{r+1}^j$ and $j\le k-r-1$. We seek the state immediately before the last update in the form
\[
(q,\e^{-\Delta},Y).
\]
The exact equations are
\begin{align}
\log q^+
&=
\log q-K_{r-1}h_B(q)+b_rh_B(\e^{-\Delta}),
\label{eq:newpulseq}\\
E
&=
h_B(Y)-\log Y-\varepsilon h_B(\e^{-\Delta}),
\label{eq:newgateY}\\
\Delta
&=
d^++b_r^\top h_B(q)-h_B(\e^{-\Delta})+\varepsilon h_B(Y).
\label{eq:newgateDelta}
\end{align}
For each sufficiently large $\Delta$, the first equation has the selected solution $q(\Delta)$ constructed above. It lies in a fixed compact subset of the interior of $Q_r^{j+1}$.

Now fix $\Delta$ and solve \eqref{eq:newgateY}. Define
\[
H_B(Y)=h_B(Y)-\log Y.
\]
For $Y\in[E/2,2E]$ and $E\le B^{1/4}$,
\[
H_B'(Y)
=
h_B'(Y)-\frac1Y
\ge \frac12
\]
when $E$ is large. At $Y=E/2$, we have $H_B(Y)<E$. At $Y=2E$, \eqref{eq:hbasic} gives
\[
h_B(2E)
\ge
2E-\frac{4E^2}{B},
\]
so $H_B(2E)>E+1$ for large $E$. Since the perturbation $\varepsilon h_B(\e^{-\Delta})$ is smaller than one, \eqref{eq:newgateY} has a unique root in this interval. Denote it by $Y(\Delta)$.

Substituting $Y=E$ into \eqref{eq:newgateY} leaves a residual of size $O(\log E)$. Since the derivative is at least $1/2$,
\begin{equation}
Y(\Delta)=E+O(\log E).
\label{eq:Yestimate}
\end{equation}
Differentiating \eqref{eq:newgateY} with respect to $\Delta$ gives
\[
H_B'(Y)Y'(\Delta)
=
-\varepsilon t_B(\e^{-\Delta}).
\]
Therefore
\begin{equation}
|Y'(\Delta)|\le C\e^{-\Delta}.
\label{eq:Yderivative}
\end{equation}

Substitute $q(\Delta)$ and $Y(\Delta)$ into the right hand side of \eqref{eq:newgateDelta}. Since $d^+$ and the bounded profile stay in fixed compact sets, and since \eqref{eq:Yestimate} holds, this right hand side equals
\[
\varepsilon E+O(\log E).
\]
Its derivative with respect to $\Delta$ is $O(\e^{-\Delta})$ by \eqref{eq:qderivative} and \eqref{eq:Yderivative}. Hence, for large $E$, it maps
\[
[\varepsilon E/2,2\varepsilon E]
\]
into itself and has derivative at most $1/2$ in magnitude. The contraction theorem gives a fixed point. Thus the exact last update is solved and
\begin{equation}
Y=E+O(\log E),
\qquad
\Delta=\varepsilon E+O(\log E).
\label{eq:Deltaestimate}
\end{equation}
After increasing the lower threshold on $E$, we may also assume
\[
\Delta\le \frac{E}{2}
\qquad\text{and}\qquad
\Delta\le B^{1/4}.
\]

\paragraph{Constructing the old orbit ending at depth $\Delta$.}
Starting from the exact endpoint $(q,\e^{-\Delta})$, repeatedly apply level $r$ episodes backward. The first terminal profile is in $Q_r^{j+1}$. Every earlier episode begins from a bounded profile in the fixed compact interior of $Q_r^0$. Hence every invocation satisfies the induction hypothesis.

Write $E_0=\Delta$. If $E_{\ell+1}$ is the initial depth of the episode whose terminal depth is $E_\ell$, then \eqref{eq:inddepth} gives
\begin{equation}
\frac{E_\ell}{\Lambda_r}
\le
E_{\ell+1}
\le
\frac{E_\ell}{2}.
\label{eq:depthchain}
\end{equation}
Choose $E$ large enough that $\Delta>\Lambda_rA$. Apply an episode
whenever the current depth exceeds $\Lambda_rA$, and stop at the first
depth at most $\Lambda_rA$. Every application then has terminal depth
greater than $A>E_r^*$. At the last application, the lower bound in
\eqref{eq:depthchain} keeps the resulting depth greater than $A$.
The upper bound halves the depth at each application, so the process
stops after $O(\log\Delta)$ episodes with depth in $(A,\Lambda_rA]$.

Reading the concatenation in forward time gives an exact old orbit ending at $(q,\e^{-\Delta})$. Let $S_g$ be the total coefficient mass of its current gate. Since $E_\ell\le2^{-\ell}\Delta$, the mass estimates in the induction hypothesis give
\begin{equation}
c_r\Delta
\le
S_g
\le
2C_r\Delta.
\label{eq:oldgatemass}
\end{equation}
The same geometric sum gives total coefficient mass at most $2C_r\Delta$ and maximum coordinate at most $C\Delta$. The last level $r$ episode in forward time also gives
\begin{equation}
h_B(p_{\mathrm{old\ gate},T-1})
\ge
c_r\Delta.
\label{eq:oldgatepulse}
\end{equation}

We now compute the length. There are at most $C\log\Delta$ episodes, and each has length at most $C_r(\log\Delta)^{r-1}$. Hence
\[
T\le C(\log\Delta)^r.
\]
For the lower bound, \eqref{eq:depthchain} gives
\[
E_\ell\ge\Delta\Lambda_r^{-\ell}.
\]
For every integer
\[
0\le\ell\le\frac{\log\Delta}{2\log\Lambda_r},
\]
we have $E_\ell\ge\sqrt\Delta$. These episodes all occur when $\Delta$ is large. Each one has length at least
\[
c_r\left(\frac12\log\Delta\right)^{r-1}.
\]
There are at least a fixed positive multiple of $\log\Delta$ such episodes. Therefore
\begin{equation}
T=\Theta((\log\Delta)^r).
\label{eq:oldconcatenatedlength}
\end{equation}

\paragraph{Adding the new gate along the old orbit.}
Apply Lemma~\ref{lem:appendixextension} to the old orbit with coupling vector
\[
b=\varepsilon e_{r+1}
\]
and terminal new coordinate $Y$ from \eqref{eq:Deltaestimate}. By \eqref{eq:oldgatepulse},
\[
b^\top h_B(p_{T-1})
\ge
\varepsilon c_r\Delta.
\]
Also $|\log Y|=O(\log E)=O(\log\Delta)$ because $\Delta=\Theta(E)$. The length, mass, maximum coordinate, and coordinate range assumptions of Lemma~\ref{lem:appendixextension} have all been verified. Thus the enlarged orbit exists exactly. The new gate is exponentially small before its final activation, and every old logarithmic coordinate changes by at most $\exp(-c\Delta)$.

Let $D_{\mathrm{new}}$ be the initial depth of the new gate. Formula \eqref{eq:initialgatedepth} and \eqref{eq:oldgatemass} give
\[
D_{\mathrm{new}}
=
\varepsilon S_g-\log Y+O(\exp(-c\Delta)).
\]
By \eqref{eq:Deltaestimate}, for large $E$,
\[
\frac{\varepsilon E}{2}
\le
\Delta
\le
2\varepsilon E.
\]
Therefore
\[
D_{\mathrm{new}}
\ge
\frac{c_r\varepsilon^2}{2}E-O(\log E)
\]
and
\[
D_{\mathrm{new}}
\le
4C_r\varepsilon^2E+O(\log E).
\]
The coupling $\varepsilon$ was chosen after $C_r$ was fixed. By \eqref{eq:epschoice}, the upper coefficient is smaller than $1/2$. After increasing the lower threshold on $E$, the logarithmic term is absorbed into the linear term. Hence there is a constant $\Lambda_{r+1}\ge4$ such that
\begin{equation}
\frac{E}{\Lambda_{r+1}}
\le
D_{\mathrm{new}}
\le
\frac{E}{2}.
\label{eq:newdepthratio}
\end{equation}

Since $\log\Delta=\log E+O(1)$, \eqref{eq:oldconcatenatedlength} gives
length $\Theta((\log E)^r)$ after adding the last update. Its gate
coefficient is $h_B(Y)=\Theta(E)$. All earlier new gate coefficients
are exponentially small. The old total mass is $O(\Delta)=O(E)$, so
the total mass at the new level is $O(E)$. The maximum coordinate is
also $O(E)$ by the old coordinate bound and $Y=E+O(\log E)$.

The old initial bounded profile lies a fixed positive distance inside
$Q_r^0$. Its initial gate depth lies in $(A,\Lambda_rA]$, whose closure
is strictly inside $I_0$. The logarithmic correction therefore keeps
the new initial bounded profile in a fixed compact subset of the
interior of $Q_{r+1}^0$. The coordinate ranges are preserved by the
strict bounds chosen above. Finally, append the exact update from
\eqref{eq:newpulseq} through \eqref{eq:newgateDelta}. Its input is the
uncorrected terminal state of the extended orbit, so the two pieces
meet exactly. This adds one update and proves \eqref{eq:indmass} at
level $r+1$. All constants and compact sets are fixed before its new
lower threshold on $E$ is chosen. This completes the induction and
proves Lemma~\ref{lem:sketch-hierarchy} in its precise form
\eqref{eq:inddepth} through \eqref{eq:indmass}.

\paragraph{Step 4. Concatenating the final level.}
For $k\ge2$, take the matrix $K_k$. Choose a fixed bounded terminal profile and set the terminal gate depth to
\[
E_0=B^{1/4}.
\]
Repeatedly apply the level $k$ episode backward until the depth reaches a fixed constant scale. By \eqref{eq:inddepth}, the depths decrease by fixed factors. The same calculation used in \eqref{eq:oldconcatenatedlength}, now with episode length $\Theta((\log E)^{k-1})$, gives a total length
\begin{equation}
T=\Theta((\log B)^k).
\label{eq:finallength}
\end{equation}
The geometric sum of the terminal depths and \eqref{eq:indmass} give
\begin{equation}\label{eq:finalorbitbounds}
\sum_{t<T}\sum_i h_B(p_{i,t})=O(B^{1/4}),
\qquad \max_{0\le t\le T,\,i}p_{i,t}=O(B^{1/4}).
\end{equation}
The starting depth stays between fixed positive bounds, and the bounded
profile belongs to a fixed compact subset of the positive orthant.
Their product gives the fixed compact set $\mathcal P$ in the
proposition, independently of $B$.

For $k=1$, take the pair construction from Step 1 with $X=B^{1/4}$. Equations \eqref{eq:pairlength} and \eqref{eq:pairmass} give the same conclusions with length $\Theta(\log B)$.

\paragraph{Step 5. A late strict increase.}
We use the scaled loss $L_B$ from \eqref{eq:scaledloss}.

Assume first that $k\ge2$. Consider the final level $k$ episode in the concatenation above. Its terminal depth is $E=B^{1/4}$. Let $t_0$ be its starting time. The episode itself has length $O((\log B)^{k-1})$, while the entire orbit has length $\Theta((\log B)^k)$. Therefore
\begin{equation}
t_0
=
T-O((\log B)^{k-1})
=
\Theta((\log B)^k).
\label{eq:t0late}
\end{equation}
At time $t_0$, the bounded profile stays in a fixed compact set and the gate is exponentially small in $E$. Hence
\begin{equation}
L_B(p_{t_0})\le C.
\label{eq:boundedlossstart}
\end{equation}
During the same episode, the final gate produces a coefficient at least $cE$. Since $h_B(x)\le x$, at some later time $t_1$ the gate coordinate itself satisfies
\[
p_{\mathrm{gate},t_1}\ge cE.
\]
Thus
\[
L_B(p_{t_1})
\ge
B\log\left(1+\frac{cB^{1/4}}{B}\right).
\]
For small $z\ge0$, $\log(1+z)\ge z/2$. Here $z=cB^{-3/4}$, so for large $B$,
\begin{equation}
L_B(p_{t_1})\ge c'B^{1/4}.
\label{eq:largelosspulse}
\end{equation}
Equations \eqref{eq:boundedlossstart} and \eqref{eq:largelosspulse} show that the scaled loss is much larger at $t_1$ than at $t_0$. Telescoping the difference
\[
L_B(p_{t_1})-L_B(p_{t_0})
=
\sum_{j=t_0}^{t_1-1}\bigl(L_B(p_{j+1})-L_B(p_j)\bigr)
\]
shows that at least one update with $j\ge t_0$ has a strict loss increase. By \eqref{eq:t0late}, this update occurs at time $\Theta((\log B)^k)$.

For $k=1$, use the final update of the pair orbit. The target is
\[
X=B^{1/4},
\qquad
Y=\frac{X}{a}\exp(-ah_B(X)).
\]
Its predecessor satisfies
\[
v=aX+O(1),
\qquad
u\le \exp(-cX).
\]
Since $X=o(B)$,
\[
B\log(1+x/B)=x+O(x^2/B)
\]
uniformly for $0\le x\le X$. Therefore
\[
L_B(X,Y)-L_B(u,v)
=
(1-a)X+O(1)+O(X^2/B)>0
\]
for large $B$. By \eqref{eq:pairlength}, this final update has index $\Theta(\log B)$. The proposition follows.
\end{proof}

\subsection{Realization by gradient descent from the origin}
\label{sub:lowerrealization}

The auxiliary orbit starts from a positive compact set, while the original problem requires gradient descent to start from $w_0=0$. We now add one extra direction and use the first gradient update to place the auxiliary coordinates at the required initial state.

\begin{proposition}
\label{prop:appendixrealization}
Let $k=\min\{n-2,d-2\}\ge1$. There are constants
$\bar\gamma_k,c_k>0$ such that, for every sufficiently large $\eta$,
some $D\in\mathcal D_{n,d}(\bar\gamma_k)$ satisfies
\[
\sigma_\eta(D)\ge c_k(1+\log\eta)^k.
\]
The stepsize threshold may depend on $n$ and $k$.
\end{proposition}

\begin{proof}
Let
\[
m=k+1,
\qquad
s=n-m.
\]
Since $k\le n-2$, we have $s\ge1$. Let $K$ be the matrix from
Proposition~\ref{prop:appendixorbit}. It is positive definite and has
unit diagonal. Therefore there are linearly independent vectors
$\widetilde v_1,\ldots,\widetilde v_m\in\R^m$ with Gram matrix $K$.
For example, if $K^{1/2}$ is its positive definite square root, take
\[
\widetilde v_i=K^{1/2}e_i.
\]
Then
\[
\ip{\widetilde v_i}{\widetilde v_j}=K_{ij},
\qquad
\norm{\widetilde v_i}=1.
\]

Choose a fixed $\rho>0$ sufficiently small and set
\[
v_i=\rho\widetilde v_i.
\]
Let
\[
S=\sum_{i=1}^m v_i.
\]
We choose $\rho$ small enough that
\begin{equation}
\norm{S}\le \frac18.
\label{eq:Ssmall}
\end{equation}
The Gram matrix of the $v_i$ is $\rho^2K$.

Recall that the original coefficient scale is $B=\eta/n$. Define
\[
\widehat B=\rho^2B.
\]
Construct the orbit in the proof of
Proposition~\ref{prop:appendixorbit}, with its length and mass estimates
\eqref{eq:finallength} and \eqref{eq:finalorbitbounds}, using parameter
$\widehat B$. For $k=1$, use \eqref{eq:pairlength} and
\eqref{eq:pairmass} instead. Denote the orbit by $\widehat p_t$ and
rescale its coordinates by
\[
p_t^*=\frac{\widehat p_t}{\rho^2}.
\]
Because
\[
h_{\widehat B}(\widehat p)
=
\frac{\widehat p}{1+\widehat p/(\rho^2B)}
=
\rho^2h_B(\widehat p/\rho^2),
\]
we have
\[
p_{t+1}^*
=
p_t^*\odot
\exp\bigl(-\rho^2Kh_B(p_t^*)\bigr).
\]
Thus $p_t^*$ is an exact isolated coefficient orbit for the vectors $v_i$ at the original scale $B$.

The scaling also preserves the loss structure up to a fixed constant. Indeed,
\[
L_{\widehat B}(\widehat p)
=
\widehat B\sum_i\log\left(1+\frac{\widehat p_i}{\widehat B}\right)
=
\rho^2B\sum_i\log\left(1+\frac{p_i^*}{B}\right)
=
\rho^2L_B(p^*).
\]
Since $\rho$ is fixed, a bounded loss state for the $\widehat B$ orbit remains bounded for the rescaled orbit, and a pulse of order $\widehat B^{1/4}$ becomes a pulse of order $B^{1/4}$ up to a fixed multiplicative constant.

Its length is $\Theta((\log B)^k)$. Its total coefficient mass and its largest coordinate are $O(B^{1/4})$. Its initial point remains in a fixed compact subset of the positive orthant. In particular,
\begin{equation}
\mu_i:=\log\frac{B}{p_{i,0}^*}
=
\log B+O(1).
\label{eq:initialmargins}
\end{equation}

Let $V$ be the $m\times m$ matrix whose columns are the $v_i$. Since these vectors are linearly independent, $V$ is invertible. Choose $u_B\in\R^m$ so that
\begin{equation}
V^\top u_B
=
\frac{2}{B}(\mu_1,\ldots,\mu_m)^\top.
\label{eq:uBdef}
\end{equation}
By \eqref{eq:initialmargins},
\begin{equation}
\norm{u_B}
\le
\frac{C\log B}{B}.
\label{eq:uBsmall}
\end{equation}

We now define the actual data in $\R^{m+1}$. For $1\le i\le m$, let
\[
x_i=(v_i,0).
\]
Use the same additional vector for all remaining samples:
\begin{equation}
y_B
=
\left(
\frac{u_B-S}{s},
\frac12
\right),
\qquad
x_{m+1}=\cdots=x_n=y_B.
\label{eq:setupvector}
\end{equation}
Since $m+1=k+2\le d$, we append zero coordinates if necessary.

We first check the norm condition. By \eqref{eq:Ssmall} and \eqref{eq:uBsmall}, for large $B$,
\[
\norm{u_B-S}\le \frac14.
\]
Therefore
\[
\norm{y_B}^2
=
\frac{\norm{u_B-S}^2}{s^2}+\frac14
\le
\frac1{16}+\frac14<1.
\]
The vectors $x_i$ with $i\le m$ also have norm at most one because $\rho\le1$.

Next we construct a separator with a margin independent of $B$. Let
\[
v=(V^\top)^{-1}\mathbf 1.
\]
Then
\[
\ip{v}{v_i}=1
\]
for every $i$. Set
\[
w_{\mathrm{aux}}=\frac{v}{\norm v},
\qquad
\gamma_{\mathrm{aux}}=\frac1{\norm v}.
\]
Thus
\[
\ip{w_{\mathrm{aux}}}{v_i}=\gamma_{\mathrm{aux}}.
\]
Define
\[
w^\star
=
\left(
\frac12w_{\mathrm{aux}},
\frac{\sqrt3}{2}
\right).
\]
This is a unit vector. For $i\le m$,
\[
\ip{w^\star}{x_i}
=
\frac{\gamma_{\mathrm{aux}}}{2}.
\]
For the copied vector $y_B$,
\[
\ip{w^\star}{y_B}
=
\frac{1}{2s}\ip{w_{\mathrm{aux}}}{u_B-S}
+
\frac{\sqrt3}{4}.
\]
By taking $\rho$ small enough first, and then $B$ large enough, the absolute value of the first term is at most $1/8$. Hence
\[
\ip{w^\star}{y_B}
\ge
-\frac18+\frac{\sqrt3}{4}
>
\frac14.
\]
Therefore every sample has margin at least
\begin{equation}
\bar\gamma_k
=
\min\left\{\frac{\gamma_{\mathrm{aux}}}{2},\frac14\right\}>0.
\label{eq:gammak}
\end{equation}
This constant depends only on the fixed auxiliary construction.

We now check the first gradient update. At $w_0=0$, every logistic coefficient equals $B/2$. Therefore
\[
\begin{aligned}
w_1
&=
\frac{B}{2}
\left(
\sum_{i=1}^m(v_i,0)+sy_B
\right)\\
&=
\frac{B}{2}
\left(
S+u_B-S,
\frac{s}{2}
\right)\\
&=
\left(
\frac{B}{2}u_B,
\frac{Bs}{4}
\right).
\end{aligned}
\]
For $i\le m$, \eqref{eq:uBdef} gives
\[
\ip{w_1}{x_i}
=
\frac{B}{2}\ip{u_B}{v_i}
=
\mu_i.
\]
Hence
\[
B\exp\bigl(-\ip{w_1}{x_i}\bigr)
=
p_{i,0}^*.
\]
Thus the auxiliary coefficient state is hit exactly after the first update.

The copied setup samples must remain negligible afterwards. Their margin at $w_1$ is
\[
\begin{aligned}
\ip{w_1}{y_B}
&=
\frac{B}{2s}\ip{u_B}{u_B-S}
+
\frac{Bs}{8}.
\end{aligned}
\]
By \eqref{eq:uBsmall}, the first term is $O(\log B)$ uniformly over $s\ge1$. Therefore
\begin{equation}
\ip{w_1}{y_B}
\ge
\frac{B}{16}
\label{eq:setupinitialmargin}
\end{equation}
for all sufficiently large $B$.

Let
\[
z_t=\log p_t^*
\]
be the logarithmic state of the isolated auxiliary orbit. Let $\widetilde z_t$ be the true logarithmic auxiliary coefficient state at the iterate $w_{1+t}$. Define
\[
M_t=\ip{w_{1+t}}{y_B},
\qquad
g_t=B\e^{-M_t},
\qquad
\epsilon_t=\norm{\widetilde z_t-z_t}.
\]
We have $\epsilon_0=0$.

Assume temporarily that $\epsilon_t\le1$ on an initial time interval. Then each true auxiliary coefficient differs from its reference value by at most a factor $\e$. Since $h_B(\e x)\le \e h_B(x)$,
\begin{equation}
\sum_{t<U}\sum_{i=1}^m h_B(\e^{\widetilde z_{i,t}})
\le
C_kB^{1/4}
\label{eq:trueauxmass}
\end{equation}
for every prefix of length $U$ inside this interval.

The setup margin evolves according to
\[
M_{t+1}-M_t
=
\sum_{i=1}^m
h_B(\e^{\widetilde z_{i,t}})
\ip{x_i}{y_B}
+
s h_B(g_t)\norm{y_B}^2.
\]
The last term is nonnegative. Also $|\ip{x_i}{y_B}|\le1$. Using \eqref{eq:setupinitialmargin} and \eqref{eq:trueauxmass},
\[
M_t
\ge
\frac{B}{16}-C_kB^{1/4}
\ge
\frac{B}{32}
\]
for all sufficiently large $B$. Consequently
\begin{equation}
h_B(g_t)
\le
g_t
\le
B\e^{-B/32}.
\label{eq:setupcoeffsmall}
\end{equation}

The isolated logarithmic update is
\[
\Phi_{B,\rho^2K}(z)
=
z-\rho^2Kh_B(\e^z).
\]
The true auxiliary update contains one additional term from the $s$ copies of $y_B$. Its $i$th component has magnitude
\[
s|\ip{x_i}{y_B}|h_B(g_t).
\]
Because
\[
s\ip{x_i}{y_B}
=
\ip{v_i}{u_B-S},
\]
this factor is bounded independently of $s$. Hence the additional update vector has norm at most
\begin{equation}
C_kB\e^{-B/32}.
\label{eq:forcingerror}
\end{equation}

On the unit ball around the reference orbit, all auxiliary coordinates are $O(B^{1/4})$. Since $t_B(x)\le x$, the derivative of the isolated logarithmic map satisfies
\[
\norm{D\Phi_{B,\rho^2K}}
\le
1+C_kB^{1/4}.
\]
Combining this with \eqref{eq:forcingerror},
\begin{equation}
\epsilon_{t+1}
\le
(1+C_kB^{1/4})\epsilon_t
+
C_kB\e^{-B/32}.
\label{eq:errorrec}
\end{equation}
The auxiliary orbit has length at most $C_k(\log B)^k$. Iterating \eqref{eq:errorrec} from $\epsilon_0=0$ gives
\[
\max_{t\le T}\epsilon_t
\le
C_kBT(1+C_kB^{1/4})^T\e^{-B/32}.
\]
Taking logarithms of the prefactor shows that it contributes only
\[
O((1+\log B)^{k+1}),
\]
which is negligible compared with $B/32$. Hence
\begin{equation}
\max_{t\le T}\epsilon_t
\le
\e^{-B/64}
\label{eq:trajectoryclose}
\end{equation}
for large $B$.

This also justifies the temporary assumption $\epsilon_t\le1$. If there were a first time at which the error exceeded one, the argument above would apply to the entire preceding prefix and would imply that the error at that first exit time is at most $\e^{-B/64}<1$, a contradiction.

It remains to transfer the late loss increase. For $k\ge2$, let $t_0<t_1$ be the bounded state and the later pulse state from the auxiliary construction. We have
\[
t_0\ge c_k(\log B)^k,
\qquad
L_B(p_{t_0}^*)\le C_k,
\qquad
L_B(p_{t_1}^*)\ge c_kB^{1/4}.
\]
The logarithmic error \eqref{eq:trajectoryclose} changes the auxiliary scaled loss by $o(1)$ at the bounded state and by $o(B^{1/4})$ at the pulse state. To see this, note that
\[
\frac{\partial}{\partial\log p}
\left[B\log\left(1+\frac pB\right)\right]
=
h_B(p)
\le p,
\]
and all reference coordinates are $O(B^{1/4})$.

The copied setup samples contribute at most
\[
sB\e^{-B/32}=o(1)
\]
to the scaled loss at every time under consideration. Therefore
\[
\eta F(w_{1+t_0})\le C_k+o(1),
\qquad
\eta F(w_{1+t_1})\ge c_kB^{1/4}+o(B^{1/4}).
\]
For large $B$, the second quantity is strictly larger. Telescoping shows that some update with index at least $1+t_0$ strictly increases the true loss. Hence
\[
\sigma_\eta(D)
\ge
c_k(\log B)^k.
\]

When $k=1$, the auxiliary pair orbit has a final scaled loss increase of order $B^{1/4}$. The error estimate \eqref{eq:trajectoryclose} changes it by only $o(B^{1/4})$, and the setup samples contribute $o(1)$. Thus the same strict increase occurs for the true gradient descent trajectory at an index of order $\log B$.

Finally, $n$ is fixed and $B=\eta/n$. Therefore
\[
\log B=\log\eta-\log n
\]
and for all sufficiently large $\eta$,
\[
\log B\ge \frac14(1+\log\eta).
\]
This proves the proposition.
\end{proof}

\subsection{Completion of the lower bound}
\label{sub:lowercomplete}

It remains to handle $k=0$, which means $n=2$ or $d=2$. We only need a constant lower bound on the transition time.

Choose
\[
u=\frac14,
\qquad
v=\frac34.
\]
Let $g_n>0$ be small enough that
\[
ng_n^2\le \frac1{32}.
\]
In two dimensions, define
\[
x_1=(u,g_n),
\qquad
x_2=\cdots=x_n=(-v,g_n),
\qquad
w^\star=(0,1).
\]
After decreasing $g_n$ if necessary, all sample norms are at most one. The vector $w^\star$ separates every sample with margin $g_n$.

At the origin every logistic coefficient equals $B/2$, so
\[
w_1=\frac{B}{2}\sum_{j=1}^n x_j.
\]
For the first sample,
\[
\begin{aligned}
\ip{w_1}{x_1}
&=
\frac{B}{2}
\left(u^2-(n-1)uv+ng_n^2\right)\\
&\le
\frac{B}{2}
\left(-\frac18+\frac1{32}\right)\\
&=
-\frac{3B}{64}.
\end{aligned}
\]
Hence
\[
F(w_1)
\ge
\frac1n\log\left(1+\exp(3B/64)\right).
\]
For large $B$, this is larger than
\[
F(w_0)=\log2.
\]
Thus the first update strictly increases the loss and
\[
\sigma_\eta(D)\ge1.
\]
If $d>2$, append zero coordinates.

For $k\ge1$, Proposition~\ref{prop:appendixrealization} gives a margin $\bar\gamma_k>0$ that does not depend on $B$, on the number of copied setup samples, or on the ambient dimension. For a fixed sample size $n$, only the finitely many values
\[
k=0,1,\ldots,n-2
\]
can occur. For $n=2$, set $\gamma_2=g_2$. For $n\ge3$, define
\[
\gamma_n
=
\min\{g_n,\bar\gamma_1,\ldots,\bar\gamma_{n-2}\}>0.
\]
After taking the largest of the finitely many stepsize thresholds, the construction gives, uniformly over every $d\ge2$,
\[
\sup_{D\in\mathcal D_{n,d}(\gamma_n)}
\sigma_\eta(D)
\ge
c_n(1+\log\eta)^{\min\{n-2,d-2\}}.
\]
If $0<\gamma\le\gamma_n$, then
\[
\mathcal D_{n,d}(\gamma_n)
\subseteq
\mathcal D_{n,d}(\gamma),
\]
so the same lower bound holds for $\mathcal D_{n,d}(\gamma)$. This is the lower bound required in Theorem~\ref{thm:main}.

\section{Proof of Theorem~\ref{thm:main}}
\label{app:mainproof}

\begin{proof}
Fix $n\ge2$ and $0<\gamma\le\gamma_n$, where $\gamma_n$ is chosen in
\Cref{sub:lowercomplete}.  The lower construction gives, uniformly
for all $d\ge2$ and all sufficiently large $\eta$,
\[
\sup_{D\in\mathcal D_{n,d}(\gamma)}\sigma_\eta(D)
\ge
c_n(1+\log\eta)^{\min\{n-2,d-2\}}.
\]
On the other hand, \Cref{app:upper} proves for every fixed
$0<\gamma\le1$ that
\[
\tau_\eta(D)
\le
C_{n,\gamma}(1+\log\eta)^{\min\{n-2,d-2\}}
\]
uniformly over $D\in\mathcal D_{n,d}(\gamma)$ and $d\ge2$.
Lemma~\ref{lem:sketch-stability} gives $\sigma_\eta(D)\le\tau_\eta(D)$,
so the same expression is an upper bound for $\sigma_\eta(D)$.  Taking the
supremum over $D$ proves Theorem~\ref{thm:main}.  All constants and the
stepsize threshold depend only on $n$ and $\gamma$, not on $d$.
\end{proof}

\section{Proofs of the corollaries}
\label{app:corollaries}

\subsection{Proof of Corollary~\ref{cor:threshold}}

\begin{proof}
Fix $n\ge2$ and $0<\gamma\le\gamma_n$.
Lemma~\ref{lem:sketch-stability} gives
$\sigma_\eta(D)\le\tau_\eta(D)$ for each dataset.
Together with \eqref{eq:appendixupper}, this yields
\[
\sup_{D\in\mathcal D_{n,d}(\gamma)}\sigma_\eta(D)
\le\sup_{D\in\mathcal D_{n,d}(\gamma)}\tau_\eta(D)
\le C_{n,\gamma}(1+\log\eta)^{\min\{n-2,d-2\}}.
\]
Theorem~\ref{thm:main} gives the matching lower bound for the first
supremum. Both constants and the common stepsize threshold are
independent of $d$, as required.
\end{proof}

\subsection{Proof of Corollary~\ref{cor:optimization}}

\begin{proof}
We first record a simple decay estimate after the threshold time.  Suppose
$F(w_t)\le1/(8\eta)$ and $\eta$ is sufficiently large depending only on
$n$.  Write
\[
\ell_{i,t}=\log(1+\exp(-\ip{w_t}{x_i})),
\qquad
q_{i,t}=\frac1{1+\exp(\ip{w_t}{x_i})}
       =1-\exp(-\ell_{i,t}).
\]
Since $\ell_{i,t}\le nF(w_t)\le n/(8\eta)\le1$, the elementary bound
$1-\exp(-x)\ge x/2$ for $0\le x\le1$ gives
\begin{align}
-\ip{\nabla F(w_t)}{w^\star}
&=\frac1n\sum_i q_{i,t}\ip{w^\star}{x_i}
\ge\frac\gamma n\sum_i q_{i,t}
\ge\frac\gamma2F(w_t).
\label{eq:postgradlower}
\end{align}
Hence $\norm{\nabla F(w_t)}\ge\gamma F(w_t)/2$.

Since $\eta F(w_t)\le1/8$, we have
$1-\eta F(w_t)/2\ge15/16$. Substituting the gradient lower bound
into \eqref{eq:loglossdescent} gives
\[
\log\frac{F(w_{t+1})}{F(w_t)}
\le-\frac{15\gamma^2}{64}\eta F(w_t).
\]
All losses are positive. Exponentiating, taking reciprocals, and using
$\exp(x)\ge1+x$ gives
\begin{equation}\label{eq:postrecurrence}
\frac1{F(w_{t+1})}
\ge\frac{\exp(15\gamma^2\eta F(w_t)/64)}{F(w_t)}
\ge\frac1{F(w_t)}+\frac{15\gamma^2}{64}\eta.
\end{equation}
Lemma~\ref{lem:sketch-stability} ensures that the threshold condition
continues to hold. Sum \eqref{eq:postrecurrence} over the $s$ updates
starting at $\tau_\eta$ and discard the positive initial reciprocal.
For every integer $s\ge1$, we obtain
\begin{equation}\label{eq:postrate}
F(w_{\tau_\eta+s})
\le\frac{64}{15\gamma^2\eta s}.
\end{equation}

Now fix $n\ge3$ and $0<\gamma\le1$.  The upper bound in
\Cref{app:upper} supplies constants $C_0$ and $\eta_0$, uniform in
$d$, such that
\[
\tau_\eta(D)\le C_0(1+\log\eta)^k,
\qquad k=\min\{n-2,d-2\}\ge1.
\]
Increase $\eta_0$ to at least $\max\{\e,n/8\}$ so that the preceding
decay estimate applies. Choose
\[
a=\min\left\{1,\frac1{2^{n-1}C_0}\right\}.
\]
Choose an integer $T_0\ge1$ with
$aT_0^{1/(n-2)}\ge\log\eta_0$.
For $T\ge T_0$ and $1\le k\le n-2$, set
$\eta=\exp(aT^{1/k})$. Then
$aT^{1/k}\ge aT^{1/(n-2)}\ge\log\eta_0\ge1$, so
$\eta\ge\eta_0$ and
\[
\tau_\eta(D)
\le C_0(2aT^{1/k})^k
\le C_0 2^{n-2}a^kT
\le\frac T2.
\]
Applying \eqref{eq:postrate} with
$s=T-\tau_\eta(D)\ge T/2$ yields
\[
F(w_T)
\le
\frac{128}{15\gamma^2\eta T}
\le
\frac{C}{T}\exp(-aT^{1/k}),
\]
which proves Corollary~\ref{cor:optimization} with
$C=128/(15\gamma^2)$. The choices of $a$, $C$, and $T_0$ depend only
on $n$ and $\gamma$.
\end{proof}

\end{document}